\documentclass{article} 
\usepackage{iclr2026_conference,times}

\usepackage{amsmath,amsfonts,bm}

\def\eqref#1{equation~\ref{#1}}

\def\1{\bm{1}}

\DeclareMathAlphabet{\mathsfit}{\encodingdefault}{\sfdefault}{m}{sl}
\SetMathAlphabet{\mathsfit}{bold}{\encodingdefault}{\sfdefault}{bx}{n}

\usepackage{bbm}
\usepackage{hyperref}
\usepackage{url}
\usepackage[utf8]{inputenc} 
\usepackage[T1]{fontenc}    
\usepackage{hyperref}       
\usepackage{booktabs}       
\usepackage{amsfonts}       
\usepackage{nicefrac}       
\usepackage{microtype}      
\usepackage{xcolor}         
\usepackage{amsmath}
\usepackage{bm}
\usepackage{wrapfig}
\usepackage{comment}
\usepackage[noend,linesnumbered,ruled,vlined]{algorithm2e}
\usepackage{graphicx}
\usepackage{amssymb}
\usepackage{wrapfig}
\usepackage{multirow}
\usepackage{comment}
\usepackage{caption}
\usepackage{tabularx,colortbl,xcolor}
\usepackage{array}
\usepackage{tikz}  
\usepackage{colortbl}
\usepackage{xcolor}    
\usepackage{marginnote} 
\usepackage{makecell}
\usepackage[export]{adjustbox}
\usepackage{threeparttable}
\usepackage{xcolor}   
\usepackage{framed}   

\usepackage{tikz}
\usetikzlibrary{arrows.meta,positioning,fit,backgrounds,calc}

\definecolor{gtblue}{RGB}{30,90,170}
\definecolor{draftorange}{RGB}{210,110,20}
\definecolor{accgreen}{RGB}{40,140,70}
\definecolor{lightgray}{RGB}{120,120,120}
 
\newcommand{\pdraft}{\tilde{q}}   
\newcommand{\ptarget}{\tilde{p}}  
\newcommand{\ptil}{\pdraft}
\newcommand{\qtil}{\ptarget}
\newcommand{\TV}{\mathrm{TV}}
\newcommand{\EAL}{\mathrm{EAL}}
\newcommand{\topk}{\mathrm{top\text{-}}K}
\newcommand{\pa}{\mathrm{pa}}
\newcommand{\Ch}{\mathrm{Ch}}
\newcommand{\bigpath}{\mathrm{path}}
\newcommand{\V}{\mathcal{V}}
\newcommand{\Tree}{\mathcal{T}}

\colorlet{rqbg}{gray!20} 
\newenvironment{researchquestionbox}{%
  \MakeFramed{\advance\hsize-\width \FrameRestore}%
  \noindent
}{%
  \endMakeFramed
}

\title{Acceptance-Aware Draft Model Training for Speculative Decoding}

\author{%
Tianhua Xia$^{1}$ \quad Leshu Li$^{1}$ \quad Mugilan Ganesan$^{1,2}$ \quad Yifei Feng$^{1}$ \quad Haiyu Wang$^{1}$ \\
\textbf{Maximilian Egger}$^{3}$ \quad \textbf{Sai Qian Zhang}$^{1}$ \\
$^1$New York University \quad $^2$University of Waterloo \quad $^3$Independent Researcher
}

\iclrfinalcopy 
\begin{document}
\pagestyle{plain} 

\maketitle
\thispagestyle{plain}

\begin{abstract}
Speculative decoding accelerates large language model (LLM) inference by using a lightweight draft model to generate multiple candidate tokens that are subsequently verified by the target model in a single forward pass. The achievable speedup is largely determined by acceptance length, yet existing draft-model training methods predominantly optimize cross-entropy or Kullback-Leibler (KL) divergence as proxies for this metric. Although these objectives encourage the draft model to match the target distribution, they do not directly optimize the acceptance length that determines speculative decoding efficiency. Moreover, the acceptance mechanism differs between greedy and sampling-based decoding, making a single training objective suboptimal across decoding regimes.

In this work, we propose acceptance-length-aware training losses that directly optimize the expected number of accepted tokens within a speculative window. We first derive an expected accepted length (EAL) loss that explicitly maximizes the expected acceptance length under greedy verification. For sampling-based speculative decoding, we introduce a window total variation (WTV) loss that directly optimizes the distributional overlap between the temperature-scaled draft and target distributions while accounting for the sequential dependence of acceptance across the speculative window. We also introduce a weighted variant of WTV that reweights each window start by its probability of serving as an actual verification boundary at inference. We further show that both objectives can be combined with a Group Relative Policy Optimization (GRPO) stage that uses simulated acceptance length as a reward, yielding additional gains over the supervised objectives.

Extensive experiments across different target models, draft models, tasks, and decoding configurations demonstrate that our acceptance-length-aware losses improve acceptance length over conventional KL-based training. In particular, the WTV loss provides substantial gains under sampling-based decoding, while the EAL loss is better aligned with greedy verification. Overall, our results demonstrate that directly optimizing for acceptance and adapting the training objective to the decoding mode provides a more effective approach to speculative decoding than conventional distribution-matching objectives.

\end{abstract}

\section{Introduction}

Large language models (LLMs) have achieved remarkable performance across a wide range of applications, but their autoregressive generation process remains a major bottleneck for efficient inference. Each generated token typically requires a separate forward pass through the target model. Speculative decoding~\cite{leviathan2023fast, hu2026dream} addresses this bottleneck by introducing a lightweight draft model that predicts multiple future tokens, which are then verified by the target model in parallel. When the drafted tokens are accepted, multiple tokens can be generated with the cost of a single target-model verification step.
The effectiveness of speculative decoding therefore depends strongly on the draft model's acceptance behavior. In particular, the acceptance length, i.e., the number of consecutively accepted draft tokens in each verification step, directly determines how many tokens can be generated per target-model invocation. Existing approaches have primarily focused on improving draft-model architectures, including parallel decoding heads, autoregressive speculators, and multi-token prediction modules, while comparatively less attention has been paid to the training objective. Most existing methods rely on conventional cross-entropy or Kullback-Leibler (KL) divergence to match the target model's output distribution.

However, distribution matching is only an indirect proxy for speculative decoding performance. KL divergence and acceptance rate share the same global optimum when the draft model can perfectly reproduce the target distribution. In practice, draft models are deliberately much smaller than their target models and therefore have limited capacity. Consequently, they must trade off different aspects of distribution matching when they cannot exactly reproduce the target distribution. Optimizing KL therefore prioritizes minimizing distributional discrepancy, whereas speculative decoding ultimately requires a different objective: producing tokens that are likely to be accepted by the target model. This mismatch becomes more pronounced under limited draft-model capacity. LK Loss~\cite{samarin2026lk} similarly motivates direct acceptance optimization from this observation, showing that distribution-matching losses can lead to suboptimal acceptance-oriented solutions.

Moreover, the acceptance mechanism differs across decoding regimes. Under greedy decoding, a draft token is accepted when it matches the target model's top-1 prediction. Therefore, the relevant quantity is the probability that the draft predicts the target's greedy token, and the probability of accepting a longer sequence depends on the product of these per-position probabilities, as shown in Section~\ref{sec:eal}. In contrast, sampling-based speculative decoding follows a different criterion. When tokens are sampled from temperature-scaled distributions, a drafted token is accepted according to the standard rejection-sampling rule. The resulting per-position acceptance probability is equal to the overlap between the temperature-scaled target and draft distributions, as described in Section~\ref{sec:wtv}.

We propose two acceptance-length-aware training objectives tailored to different decoding regimes. For greedy decoding, we introduce a loss that optimizes the expected acceptance length by encouraging consecutive agreement with the target model. For sampling-based decoding, we develop a complementary objective tailored to the acceptance behavior induced by stochastic sampling. Our approach differs from conventional knowledge distillation in two key aspects. First, we optimize the actual acceptance length rather than using distributional similarity as a proxy. Second, we tailor the training objective to the underlying decoding regime. This distinction is particularly important for sampling-based decoding, where the acceptance behavior differs fundamentally from greedy verification. Moreover, acceptance-length objectives typically treat every token position equally as the start of a drafting round, whereas at inference a new round begins only after the previous verification step terminates. We therefore propose a weighted WTV that reweights each position by its probability of actually starting a drafting round. Beyond these two differentiable surrogates, we also investigate whether acceptance length can be optimized directly as a non-differentiable reward. We adapt Group Relative Policy Optimization (GRPO) to the draft-model setting by simulating greedy verification over window-level slices of a single forward pass. The resulting acceptance length, debiased against a frozen reference draft, serves as the reward signal. This RL stage is applied on top of a supervised checkpoint and is complementary to the surrogate losses.

We evaluate our objectives across multiple benchmarks and model configurations under both greedy and sampling-based speculative decoding. Our results demonstrate that direct acceptance-length optimization improves acceptance length over conventional KL-based training when the objective is matched to the decoding regime. In summary, our contributions are:

\begin{itemize}
    \item We formulate draft-model training around expected acceptance length, moving beyond conventional distribution matching to better align the training objective with the metric that determines speculative decoding efficiency.

    \item We derive an expected accepted length (EAL) loss for greedy verification and a window total variation (WTV) loss for sampling-based decoding, together with a weighted WTV variant that emphasizes positions where drafting rounds actually begin at inference.

    \item We demonstrate the effectiveness of the proposed losses across different models, tasks, and decoding regimes, showing that matching the training objective to the underlying acceptance mechanism consistently improves speculative decoding performance.
\end{itemize}

\section{Background and Related work}
\subsection{Speculative Decoding}
Speculative decoding is a widely adopted technique for accelerating autoregressive language model inference by decoupling token generation into a lightweight draft model and a more capable target model~\cite{leviathan2023fast, chen2023accelerating}. Given a context, the draft model generates multiple candidate tokens, which are then verified by the target model in parallel. Since rejected tokens are corrected using the target distribution, speculative decoding can preserve the target output distribution while reducing the number of expensive autoregressive model evaluations. The effectiveness of speculative decoding largely depends on how many drafted tokens can be accepted in each speculation step. We refer to this quantity as the acceptance length, which determines how many draft tokens are accepted per verification step and strongly influences the achievable decoding speedup. Consequently, draft-model quality for speculative decoding is better characterized by its acceptance behavior than by its language modeling capability alone.

Speculative decoding can operate under different decoding regimes. Under greedy decoding, a draft token is accepted when it matches the token selected by the target model. Under sampling-based decoding, acceptance instead follows a rejection-sampling procedure determined by the overlap between the target and draft probability distributions. Thus, although both settings share the same draft-then-verify framework, their underlying acceptance mechanisms differ fundamentally.

\subsection{Draft Model Architectures}
Early speculative decoding methods used a smaller pretrained language model as the draft model~\cite{leviathan2023fast, chen2023accelerating}. Subsequent work introduced more lightweight and integrated architectures. MEDUSA~\cite{cai2024medusa} uses multiple decoding heads for parallel token prediction, while EAGLE~\cite{li2024eagle} employs an autoregressive draft head over hidden representations, with further improvements in EAGLE-3~\cite{li2025eagle3}. DeepSeek-V3~\cite{deepseekai2024deepseekv3} incorporates Multi-Token Prediction (MTP)~\cite{gloeckle2024better} as a native speculative decoding module, while FR-Spec~\cite{zhao2025frspec} reduces draft overhead through frequency-based vocabulary restriction.
These works largely focus on how to efficiently generate draft tokens, while how to train draft models for better acceptance remains an important question.

\subsection{Training Losses for Draft Models}

Most draft-model training methods~\cite{cai2024medusa, li2024eagle, li2025eagle3} formulate draft training as a knowledge distillation problem. Given the target distribution \(p\) and draft distribution \(q\), the draft model is trained to approximate the target model, using forward KL divergence or its equivalent cross-entropy objective. However, distribution matching is an indirect objective for speculative decoding. The ultimate goal of a draft model is not to reproduce the target distribution in isolation, but to generate tokens that are likely to be accepted by the target model during speculative verification.

More recent studies have started to consider objectives that are more closely related to speculative decoding efficiency. DistillSpec~\cite{zhou2023distillspec}, for example, investigates alternative divergence measures, including reverse KL and total variation (TV) distance. In particular, TV distance has a direct connection to the acceptance probability under sampling-based speculative decoding. AdaSPEC~\cite{hu2025adaspec} further addresses the mismatch between conventional distillation objectives and speculative decoding efficiency through selective distillation.

LK Losses~\cite{samarin2026lk} further argues that KL divergence is only a proxy for acceptance optimization, as minimizing KL does not necessarily maximize acceptance. LK therefore introduces losses that directly target token acceptance rather than relying solely on distributional similarity.

Our work follows this acceptance-oriented direction but focuses on a different question: how should the training objective be designed to optimize acceptance length across different decoding regimes, from greedy to sampling-based decoding?

\subsection{Preliminaries and Notation}
\label{subsec:prelim}

We consider speculative decoding with two autoregressive models 
sharing the same vocabulary: the target model with next-token distribution 
$p_t(\cdot) \triangleq p(\cdot \mid x_{<t})$ and the draft model with next-token distribution 
$q_t(\cdot) \triangleq q(\cdot \mid x_{<t})$,
both conditioned on the same prefix $x_{<t}$. At each position, the 
draft proposes a token, which is then accepted or rejected by the 
verification criterion described below.

For inference at sampling temperature $T > 0$, the temperature-scaled 
distributions are:
\begin{equation}
\tilde{p}_t(v) \;\propto\; \exp\!\left(\log p_t(v) / T\right), 
\qquad
\tilde{q}_t(v) \;\propto\; \exp\!\left(\log q_t(v) / T\right).
\end{equation}

In the greedy limit $T \to 0^{+}$, both $\tilde{p}_t$ and $\tilde{q}_t$ concentrate on the argmax of their unscaled distributions.

Two acceptance criteria are commonly used in speculative decoding, and we refer to
them as \emph{greedy} and \emph{sampling} verification. 
Under \textbf{greedy verification}, the target selects its most likely token
$x^{\star}_{t} = \arg\max_{v \in \mathcal{V}} p_t(v)$, and the drafted token is accepted
only when it coincides with $x^{\star}_{t}$. The single-step acceptance probability is
therefore the probability mass assigned by the draft model to the target's selected token,
\begin{equation}
a^{\mathrm{greedy}}_{t} \;\triangleq\; q_t\!\left(x^{\star}_{t}\right)
\;=\; q\!\left(x^{\star}_{t} \mid x_{<t}\right),
\label{eq:acc_greedy}
\end{equation}
and a rejected position is replaced by $x^{\star}_{t}$ itself.

Under \textbf{sampling verification}~\cite{leviathan2023fast, chen2023accelerating},
the draft samples $\hat{y} \sim \tilde{q}_t(\cdot)$ and the token is accepted with
probability $\min{1,, \tilde{p}_t(\hat{y})/\tilde{q}_t(\hat{y})}$. Averaging over the draft
proposal distribution yields an acceptance probability equal to the overlap between the
two distributions,
\begin{equation}
a^{\mathrm{sample}}_{t}
 \;\triangleq\; \mathbb{E}_{\hat{y} \sim \tilde{q}_t}\!\left[\min\!\Big(1,\ \frac{\tilde{p}_t(\hat{y})}{\tilde{q}_t(\hat{y})}\Big)\right]
 \;=\; \sum_{v \in \mathcal{V}} \min\{\tilde{p}_t(v),\, \tilde{q}_t(v)\}
 \;=\; 1 - \mathrm{TV}(\tilde{p}_t,\, \tilde{q}_t),
\label{eq:acc_sample}
\end{equation}
and a rejected token is resampled from the residual distribution. When the criterion in
use is clear from the context, we drop the superscript and write $a_t$.

\section{Acceptance-Aware Loss Design}
\label{sec:loss_design}

\subsection{Speculative Window}
\begin{wraptable}{r}{0.5\textwidth}
\vspace{-12pt}
\centering
\scriptsize
\setlength{\tabcolsep}{4pt}
\renewcommand{\cellalign}{cc}
\caption{Organization of Section~\ref{sec:loss_design}.}
\label{tab:loss_overview}
\begin{tabular}{ccc}
\toprule
 & \textbf{Chain} & \textbf{Tree} \\
\midrule
\textbf{Uniform}
 & \makecell{\S\ref{sec:chain_eal} Chain EAL \\ \S\ref{sec:wtv} Window TV}
 & \makecell{\S\ref{sec:tree_eal} Tree EAL} \\
\midrule
\textbf{Weighted}
 & \makecell{\S\ref{sec:weighted_wtv} Weighted WTV}
 & \makecell{\S\ref{sec:weighted_tree} Weighted Tree EAL \\ \S\ref{sec:temporal_tree} Temporal Tree EAL} \\
\bottomrule
\end{tabular}
\vspace{-8pt}
\end{wraptable}
A speculative window of size $\gamma$ starting at position $t$ covers positions
$t, t+1, \ldots, t+\gamma-1$; we index positions inside the window by the offset
$j = 0, 1, \ldots, \gamma-1$. Let $L_t \in \{0, 1, \ldots, \gamma\}$ denote the number of
consecutively accepted draft tokens in the window. Verification halts at the first rejection,
so the first $k$ tokens are accepted only if each of them is accepted. We write $a_{t+j}$ for
the conditional acceptance rate of the $(j+1)$-th drafted token under the criterion in use,
given that the window starts at $t$ and the preceding $j$ drafted tokens are accepted,
$a_{t+j} \triangleq \mathbb{P}(L_t \geq j+1 \mid L_t \geq j)$. By the chain rule of probability,
\begin{equation}
\mathbb{P}(L_t \geq k) \;=\; \prod_{j=0}^{k-1} a_{t+j},
\qquad
\mathbb{E}[L_t] \;=\; \sum_{k=1}^{\gamma} \mathbb{P}(L_t \geq k)
              \;=\; \sum_{k=1}^{\gamma} \prod_{j=0}^{k-1} a_{t+j},
\label{eq:eal_generic}
\end{equation}
where the second equality is the tail-sum identity for nonnegative integer-valued random
variables. Under teacher forcing, each $a_{t+j}$ is given by the per-position acceptance
probability of Eq.~\ref{eq:acc_greedy} or Eq.~\ref{eq:acc_sample}. Equation~\ref{eq:eal_generic} is the common template for all
objectives in this section: Section~\ref{sec:eal} instantiates it with
$a_{t} = a^{\mathrm{greedy}}_{t}$ for chain and tree drafting, Section~\ref{sec:wtv} with
$a_{t} = a^{\mathrm{sample}}_{t}$, and Section~\ref{sec:weighted} reweights the window
starts according to how speculative decoding actually visits them. We write $N$ for the
length of a training sequence and $\mathcal{W}$ for its set of window start positions.

\subsection{Expected Accepted Length Loss}
\label{sec:eal}

We first derive the EAL loss for chain drafting under greedy verification, and then extend it
to tree drafting, where several candidate paths are verified in parallel.

\subsubsection{Chain Expected Accepted Length}
\label{sec:chain_eal}

Under greedy verification, a drafted token at position $t$ is accepted if and only if it
equals the target's own choice $x^{\star}_{t} = \arg\max_{v \in \mathcal{V}} p_t(v)$. Because
the training sequences are produced by the frozen target under greedy decoding,
$x^{\star}_{t}$ coincides with the ground-truth token at that position, so the per-position
acceptance probability of Eq.~\ref{eq:acc_greedy} is the probability the draft assigns to the
ground-truth continuation. Throughout this section we abbreviate
$a_{t} \triangleq a^{\mathrm{greedy}}_{t}$.

For a window of size $\gamma$ starting at position $t$, greedy verification accepts the first
$\ell$ drafted tokens and (if $\ell < \gamma$) rejects the $(\ell+1)$-th, so the distribution
of the accepted length $L_t$ is:
\begin{align}
\mathbb{P}(L_t = 0) &\;=\; 1 - a_{t}, \\
\mathbb{P}(L_t = \ell) &\;=\; \left(\prod_{j=0}^{\ell-1} a_{t+j}\right)\left(1 - a_{t+\ell}\right),
\quad \text{for } 1 \leq \ell \leq \gamma - 1, \\
\mathbb{P}(L_t = \gamma) &\;=\; \prod_{j=0}^{\gamma-1} a_{t+j},
\end{align}
where the last case has no rejection term because the window is exhausted before a rejection
occurs. The expected accepted length is:
\begin{equation}
\mathbb{E}[L_t]
\;=\; \sum_{\ell=0}^{\gamma} \ell \cdot \mathbb{P}(L_t = \ell)
\;=\; \sum_{\ell=1}^{\gamma-1} \ell \left(\prod_{j=0}^{\ell-1} a_{t+j}\right)\left(1 - a_{t+\ell}\right)
  \;+\; \gamma \prod_{j=0}^{\gamma-1} a_{t+j},
\label{eq:eal_full}
\end{equation}
where the $\ell = 0$ term vanishes. Eq.~\ref{eq:eal_full} telescopes into the tail-sum
form of Eq.~\ref{eq:eal_generic}, giving the more compact expression
\begin{equation}
\mathbb{E}[L_t] \;=\; \sum_{k=1}^{\gamma} \prod_{j=0}^{k-1} a_{t+j}.
\label{eq:eal_compact}
\end{equation}
As an example, for a window of size $\gamma = 3$, Eq.~\ref{eq:eal_full} reads
\begin{equation}
\mathbb{E}[L_t] \;=\; 1 \cdot a_{t}(1 - a_{t+1})
   \;+\; 2 \cdot a_{t} a_{t+1} (1 - a_{t+2})
   \;+\; 3 \cdot a_{t} a_{t+1} a_{t+2},
\end{equation}
which equals
\begin{equation}
\mathbb{E}[L_t] \;=\; a_{t} \;+\; a_{t} a_{t+1} \;+\; a_{t} a_{t+1} a_{t+2}.
\end{equation}

The expected accepted length is a quantity to be \emph{maximized}, so the corresponding
training loss is its negation. The per-window EAL loss and the sequence-level objective are
\begin{equation}
\mathcal{L}^{(t)}_{\mathrm{EAL}}(\theta)
 \;\triangleq\; -\sum_{k=1}^{\gamma} \prod_{j=0}^{k-1} a^{\mathrm{greedy}}_{t+j},
\qquad
\mathcal{L}_{\mathrm{EAL}}(\theta)
 \;=\; \frac{1}{|\mathcal{W}|} \sum_{t \in \mathcal{W}} \mathcal{L}^{(t)}_{\mathrm{EAL}}(\theta),
\label{eq:loss_eal}
\end{equation}
where $\theta$ denotes the draft model parameters and the target distributions are frozen.

\subsubsection{Greedy Tree Expected Accepted Length}
\label{sec:tree_eal}

\begin{figure}[t]
  \centering
  \includegraphics[width=\linewidth]{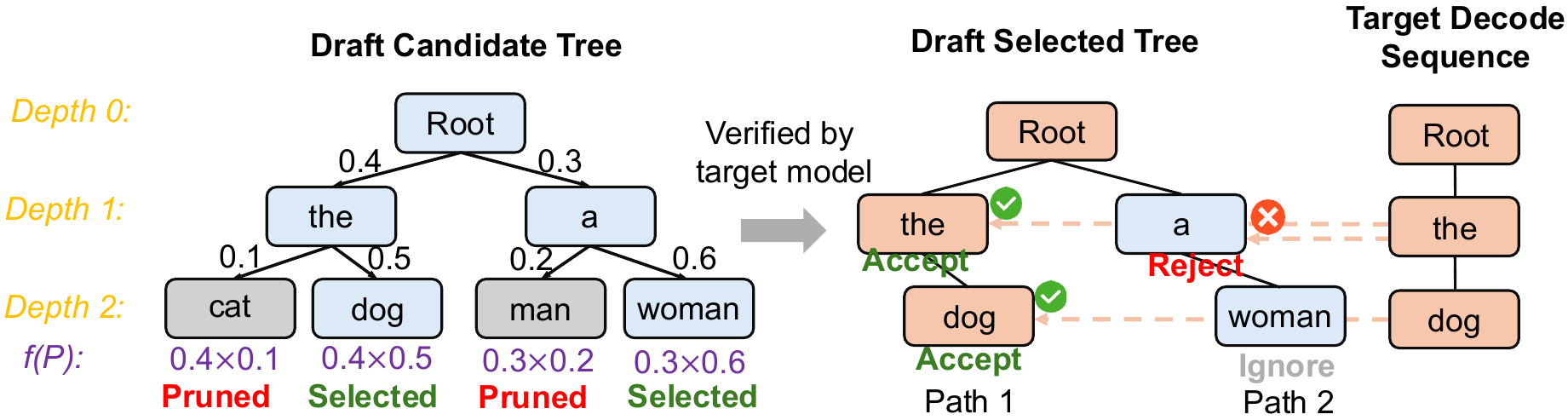}
  \caption{Greedy tree verification ($K=2$, $\gamma=2$). Left: candidate paths are scored by
  $f(P)$ and the top-$K$ are selected. Right: the selected paths are verified against the
  target greedy sequence.}
  \label{fig:tree_paths}
\end{figure}

We extend the chain-based EAL objective to tree-based speculative
decoding under greedy target verification. The draft model constructs
a tree of maximum depth $\gamma$ in two stages.

\paragraph{(i) Layer-wise expansion.}
We construct a draft tree layer by layer, where the root
represents the training prefix and has depth $0$.
Each non-root node contains a draft token and a probability
distribution over the vocabulary $\mathcal{V}$.
The token is selected from its parent's distribution,
while the node's own distribution is used to expand
its children.
At each depth, the draft model expands each node
by selecting the highest-probability tokens from its
distribution, where the number of selected tokens is
determined by the predefined tree structure.
Each edge connects a parent node to its child, with
the edge weight given by the draft probability of
the child token.

Let $P$ denote the path from the root of the draft tree to a
node at depth $j$.
We write $P=(v_1^*,\ldots,v_j^*)$, where
$\mathcal V$ is the vocabulary and $v_i^*\in\mathcal V$ is the
draft token at depth $i$ on $P$, for $i\in\{1,\ldots,j\}$.
Thus, $v_j^*$ is the token at the node.

We score each path by multiplying the edge weights along
the path:
\begin{equation}
f(P)=\prod_{i=1}^{j} q_i^P(v_i^*),
\label{eq:tree_path_score}
\end{equation}
where $q_i^P(v_i^*)$ denotes the draft probability of token
$v_i^*$ at depth $i$, conditioned on the preceding tokens
along path $P$.

\paragraph{(ii) Global pruning and re-ranking.}
After Layer-wise expansion, we rank all root-to-leaf paths of depth $\gamma$ in the generated candidate tree according to their cumulative draft probability scores $f(P)$, and retain the top-$K$ paths to form the pruned tree. We denote the $k$-th retained path by
\begin{equation}
P_k = (v_{k,1}^*, \ldots, v_{k,\gamma}^*),
\quad k \in \{1,\ldots,K\},
\label{eq:retained_path}
\end{equation}
where $v_{k,j}^*$ is the draft token at depth $j$ on path $P_k$.

\paragraph{Target Verification.}
Let $G=(g_1^*,\ldots,g_\gamma^*)$ denote the token
sequence generated by the target model using greedy decoding. At each depth $j$ from $1$
to $\gamma$, the target model selects
$g_j^*=\arg\max_{v\in\mathcal V} p_j(v)$, where $p_j(v)$ denotes
the target model's probability of token $v$ at depth $j$.

Recall that $L_t$ denotes the accepted length of the window starting
at position $t$. Let $L_t(P_k)$ denote the accepted length of path
$P_k$ starting at position $t$.
Starting at depth $1$, we compare the draft token $v_{k,j}^*$
with the target greedy token $g_j^*$ at each depth $j$.
If the first mismatch occurs at depth $j$, then $L_t(P_k)=j-1$.
If all $\gamma$ tokens match, then 
$L_t(P_k)=\gamma$.
The accepted length of the tree is
\begin{equation}
L_{\mathrm{Tree}}^{t} = \max_{1\le k\le K} L_t(P_k).
\label{eq:tree_actual_length}
\end{equation}
Figure 1 illustrates the full process with $K=2$ and $\gamma=2$: the two paths with the highest $f(P)$ are selected, while the others are pruned, and the selected paths are subsequently verified by the target model.

\begin{wrapfigure}{r}{0.42\textwidth}
  \vspace{-12pt}
  \centering
  \includegraphics[width=\linewidth]{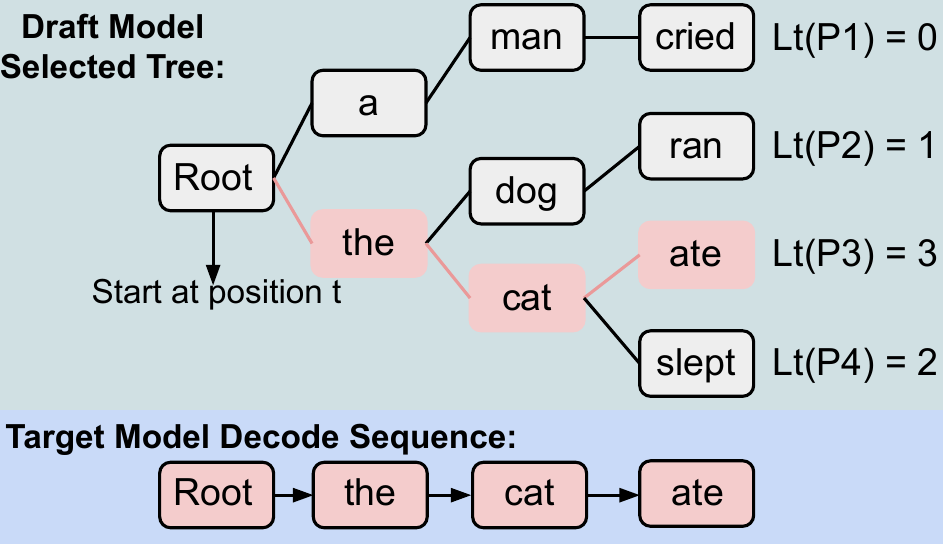}
  \vspace{-18pt}
  \captionsetup{font=footnotesize}
  \caption{Accepted length $L_t(P_k)$ of each selected path.}
  \label{fig:sd_lt}
  \vspace{-10pt}
\end{wrapfigure}
\paragraph{Differentiable Path EAL Surrogate.}
Our objective is to maximize the expected tree accepted length,
$\mathbb{E}[L_{\mathrm{Tree}}^t]$.
Following Eq.~\ref{eq:eal_compact}, the expected accepted length
of each path $P_k$ can be expressed as
\begin{equation}
\mathbb{E}[L_t(P_k)]
=
\sum_{d=1}^{\gamma}
\mathbb{P}(L_t(P_k)\ge d).
\label{eq:tree_path_expected_length}
\end{equation}
By the chain rule of conditional probability,
\begin{equation}
\mathbb{P}(L_t(P_k)\ge d)
=
\prod_{j=1}^{d}
\mathbb{P}\left(
L_t(P_k)\ge j
\mid
L_t(P_k)\ge j-1
\right).
\label{eq:tree_path_conditional}
\end{equation}
Thus,
\begin{equation}
\mathbb{E}[L_t(P_k)]
=
\sum_{d=1}^{\gamma}
\prod_{j=1}^{d}a_j(k),
\label{eq:tree_path_eal_exact}
\end{equation}
where $a_j(k)$ denotes the conditional acceptance probability
at depth $j$ along path $P_k$, given that the preceding $j-1$
tokens have been accepted.
Figure~\ref{fig:sd_lt} illustrates $L_t(P_k)$ for each path of a selected tree.

Under greedy target verification, the acceptance of a fixed
path is deterministic, with each acceptance decision being
either $0$ or $1$. We construct a differentiable surrogate
by truncating the EAL expression at $L_t(P_k)$:
\begin{equation}
\mathbb{E}[L_t(P_k)] =
\sum_{d=1}^{L_t(P_k)}
\prod_{j=1}^{d}a_j(k).
\label{eq:tree_path_eal_truncated}
\end{equation}

We then approximate the binary acceptance decision at depth
$j$ using the draft probability assigned to the target
greedy token:
\begin{equation}
a_j(k) \approx q_j^{P_k}(g_j^*),
\qquad j\le L_t(P_k),
\label{eq:tree_substitute}
\end{equation}
where $q_j^{P_k}(v)$ denotes the draft model's probability
of token $v$ at depth $j$ along path $P_k$.

Substituting this approximation yields
\begin{equation}
\mathbb{E}[L_t(P_k)]
\approx
\sum_{d=1}^{L_t(P_k)}
\prod_{j=1}^{d}q_j^{P_k}(g_j^*).
\label{eq:tree_path_eal}
\end{equation}

According to Eq.~\ref{eq:tree_actual_length}, the tree accepted
length is determined by the longest accepted path. Since greedy
verification is deterministic, the expectation of the maximum
equals the maximum of the expectations. Thus,
\begin{equation}
\begin{aligned}
\mathbb{E}[L_{\mathrm{Tree}}^t]
&=
\mathbb{E}\left[\max_{1\le k\le K}L_t(P_k)\right]\\
&=
\max_{1\le k\le K}\mathbb{E}[L_t(P_k)]\\
&\approx
\max_{1\le k\le K}
\left(
\sum_{d=1}^{L_t(P_k)}
\prod_{j=1}^{d}q_j^{P_k}(g_j^*)
\right).
\end{aligned}
\label{eq:tree_expected_length}
\end{equation}

\paragraph{Max-path Tree EAL Loss.}
Following Eq.~\ref{eq:tree_expected_length},
we maximize the largest EAL surrogate among the retained
paths. The corresponding loss is
\begin{equation}
\mathcal{L}_{\mathrm{TreeEAL\text{-}Max}}^t(\theta)
=
-\max_{1\le k\le K}
\left(
\sum_{d=1}^{L_t(P_k)}
\prod_{j=1}^{d}q_j^{P_k}(g_j^*)
\right),
\label{eq:tree_eal_max_loss}
\end{equation}
where $\theta$ denotes the draft model parameters.
Since all paths share the same draft probabilities along
their matched prefixes, each additional accepted token
contributes a nonnegative term. Therefore, the path
with the largest accepted length achieves the maximum
surrogate.

\paragraph{Average-path Tree EAL Loss.}
To optimize all retained paths, we maximize the average
of their EAL surrogates. The corresponding loss is
\begin{equation}
\mathcal{L}_{\mathrm{TreeEAL\text{-}Avg}}^t(\theta)
=
-\frac{1}{K}\sum_{k=1}^{K}
\sum_{d=1}^{L_t(P_k)}
\prod_{j=1}^{d}q_j^{P_k}(g_j^*).
\label{eq:tree_eal_loss}
\end{equation}
Each retained path contributes to the loss according
to its accepted length.

\paragraph{Variance-Regularized Tree EAL Loss.}
To encourage balanced optimization across retained paths,
we penalize the variance of their EAL surrogates:
\begin{equation}
\begin{aligned}
\sigma_t^2
=
\frac{1}{K}\sum_{k=1}^{K}
\Bigg(
&\sum_{d=1}^{L_t(P_k)}
\prod_{j=1}^{d}q_j^{P_k}(g_j^*)\\
&-\frac{1}{K}\sum_{r=1}^{K}
\sum_{d=1}^{L_t(P_r)}
\prod_{j=1}^{d}q_j^{P_r}(g_j^*)
\Bigg)^2.
\end{aligned}
\label{eq:tree_variance}
\end{equation}
We combine this penalty with the average-path EAL loss:
\begin{equation}
\mathcal{L}_{\mathrm{TreeEAL+Var}}^t(\theta)
=
\mathcal{L}_{\mathrm{TreeEAL\text{-}Avg}}^t(\theta)
+\lambda\sigma_t^2,
\label{eq:tree_eal_variance_loss}
\end{equation}
where $\lambda\ge 0$ controls the strength of
variance regularization.

\subsection{Window TV Loss}
\label{sec:wtv}

Under sampling verification, acceptance is not decided by a single token match but by the
overlap of the two temperature-scaled distributions. By Eq.~\ref{eq:acc_sample}, the
per-position rejection rate at position $t$ equals the total variation distance between them,
\begin{equation}
r_t \;\triangleq\; \mathrm{TV}\!\left(\tilde{p}_t,\; \tilde{q}_t\right),
\qquad
a^{\mathrm{sample}}_{t} \;=\; 1 - r_t.
\label{eq:rejection_rate}
\end{equation}
Throughout this section we abbreviate $a_{t} \triangleq a^{\mathrm{sample}}_{t}$.

Substituting Eq.~\ref{eq:rejection_rate} into the window template of
Eq.~\ref{eq:eal_generic}, the expected accepted length of a window of size $\gamma$ starting
at position $t$ is:
\begin{equation}
\mathbb{E}[L_t]
\;=\; \sum_{k=1}^{\gamma} \prod_{j=0}^{k-1} a_{t+j}
\;=\; \sum_{k=1}^{\gamma} \prod_{j=0}^{k-1}
      \big(1 - \mathrm{TV}(\tilde{p}_{t+j}, \tilde{q}_{t+j})\big).
\label{eq:eal_tv}
\end{equation}
The direct surrogate is the negation of Eq.~\ref{eq:eal_tv} itself, evaluated with the draft
and target models:
\begin{equation}
\mathcal{L}^{(t)}_{\mathrm{WTV}}(\theta)
\;\triangleq\;
-\sum_{k=1}^{\gamma} \prod_{j=0}^{k-1}
\Big( 1 - \mathrm{TV}(\tilde{p}_{t+j},\, \tilde{q}_{t+j}) \Big),
\qquad
\mathcal{L}_{\mathrm{WTV}}(\theta)
\;=\; \frac{1}{|\mathcal{W}|} \sum_{t \in \mathcal{W}} \mathcal{L}^{(t)}_{\mathrm{WTV}}(\theta).
\label{eq:loss_tv}
\end{equation}
The target distributions $\tilde{p}_{t+j}$ are frozen, and gradients flow only into the draft
parameters $\theta$ through $\tilde{q}_{t+j}$; the loss is differentiable because $\mathrm{TV}$
is a piecewise-linear function of $\tilde{q}$.

\begin{wrapfigure}{r}{0.48\textwidth}
  \vspace{-12pt}
  \centering
  \includegraphics[width=\linewidth]{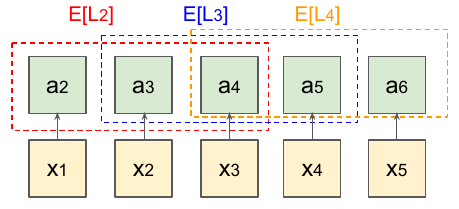}
  \vspace{-18pt}
  \caption{Illustration of the Window TV (WTV) loss with window size $\gamma = 3$.
  At each position $t$, the draft $\tilde{q}_t$ and the frozen target
  $\tilde{p}_t$ are conditioned on the same ground-truth context,
  giving $a_t = 1 - \mathrm{TV}(\tilde{p}_t, \tilde{q}_t)$. Each dashed box is
  a window starting at $t$, contributing
  $\mathbb{E}[L_t] = \sum_{k=1}^{\gamma} \prod_{j=0}^{k-1} a_{t+j}$.}
  \label{fig:wtv}
  \vspace{-10pt}
\end{wrapfigure}

Figure~\ref{fig:wtv} illustrates how $\mathcal{L}_{\mathrm{WTV}}$ is computed, reading the
diagram from the bottom up. The bottom row is the conditioning context: the draft is run once
over the target-generated sequence $x_1, x_2, \ldots$ under teacher forcing, so at every
position $t$ the draft and the frozen target share the same context. The middle row is
what each of them produces there, namely the temperature-scaled distributions $\tilde{q}_t$
and $\tilde{p}_t$; the shaded region where the two histograms overlap is exactly the quantity
$\sum_{v} \min\{\tilde{p}_t(v), \tilde{q}_t(v)\}$ that Eq.~\ref{eq:acc_sample} identifies
with the per-position acceptance probability, so the top row of the figure, the sequence of
$a_t = 1 - \mathrm{TV}(\tilde{p}_t, \tilde{q}_t)$ values, is obtained for all positions in
parallel from a single draft forward pass and a single target forward pass.

The dashed boxes are the speculative windows of Eq.~\ref{eq:eal_generic}. A window starting
at $t$ covers positions $t, \ldots, t+\gamma-1$, and because verification halts at the first
rejection, the $k$-th drafted token contributes only through the cumulative product
$\prod_{j=0}^{k-1} a_{t+j}$; summing these products over $k$ gives $\mathbb{E}[L_t]$. With
$\gamma = 3$ as drawn, the first window contributes
$\mathbb{E}[L_t] = a_{t} + a_{t}a_{t+1} + a_{t}a_{t+1}a_{t+2}$. The window then slides by one
position and the loss accumulates $-\mathbb{E}[L_t]$ over all start positions.

\subsection{Weighted Acceptance-Aware Losses}
\label{sec:weighted}

The objectives above treat every position in $\mathcal{W}$ uniformly as a draft window start, however, this approach still faces the issue of train-inference mismatch,
whereas during actual speculative decoding a new draft window begins only after the previous one
terminates. We first derive draft window-start weights that correct this mismatch and apply them
to the chain WTV loss and to the Tree EAL loss; both weightings are based on window-start \emph{probabilities}. We then introduce a temporal variant that instead uses the window starts \emph{realized} by actual tree speculative decoding.

\subsubsection{ draft Window-Start Weights}
\label{sec:start_weights}

\begin{wrapfigure}{r}{0.5\linewidth}
  \vspace{-10pt}
  \centering
  \includegraphics[width=\linewidth]{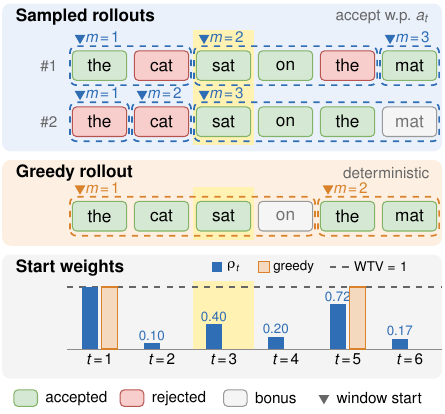}
  \vspace{-14pt}
  \caption{Window starts on one sequence ($\gamma=3$) under sampled and
  greedy rollouts, and the resulting start weights.}
  \label{fig:rollout_starts}
  \vspace{-10pt}
\end{wrapfigure}

Consider a training sequence of length $N$, generated by the target
model, whose tokens serve as the ground truth that the draft is trained
to reproduce, and let $M$ be the number of windows
draft rollout from it. Each draft rollout window advances the sequence by at least one token, so $1\leq M\leq N$.
We index windows by $m=1,\ldots,N$. Let $B^m_t=\{\text{window $m$ starts at position $t$}\}$,
and let $L^m_t$ be the accepted length of the $m$-th window when it starts at position $t$,
with $L^m_t=0$ for $m>M$. The window-start probability of position $t$ is
\begin{equation}
\rho_t
=
\sum_{m=1}^{N}\mathbb{P}(B^m_t),
\label{eq:start_prob}
\end{equation}

i.e., the probability that some window starts at $t$ (window starts are strictly increasing,
so the events $B^m_t$ are disjoint in $m$). Under teacher forcing, draft model are conditioned on the
ground-truth prefix $x_{<t}$ at every rollout position, where this process simulates greedy speculative decoding exactly and under speculative sampling it is an approximation. Consequently the accepted length of a
window depends only on its start position, and is
independent of the event $B^m_t$, which is determined by acceptances
before $t$ so
$\mathbb{E}[L^m_t\mid B^m_t]=\mathbb{E}[L_t]$. By the law of total expectation, the total
accepted length $L_{\mathrm{tot}}=\sum_{m=1}^{N}\sum_{t=1}^{N}\mathbbm{1}[B^m_t]\,L^m_t$ and the number of windows satisfy
\begin{equation}
\mathbb{E}[L_{\mathrm{tot}}]
\;=\;
\sum_{m=1}^{N}\sum_{t=1}^{N}\mathbb{P}(B^m_t)\,\mathbb{E}[L_t]
\;=\;
\sum_{t=1}^{N}\rho_t\,\mathbb{E}[L_t],
\qquad
\mathbb{E}[M]
\;=\;
\sum_{t=1}^{N}\rho_t.
\end{equation}

The expected accepted length per executed window is therefore
\begin{equation}
\frac{\mathbb{E}[L_{\mathrm{tot}}]}{\mathbb{E}[M]}
\;=\;
\sum_{t=1}^{N}\bar{\rho}_t\,\mathbb{E}[L_t],
\qquad
\bar{\rho}_t
\;=\;
\frac{\rho_t}{\sum_{t'=1}^{N}\rho_{t'}},
\qquad
\sum_{t=1}^{N}\bar{\rho}_t=1,
\label{eq:weighted_identity}
\end{equation}
which is a ratio of expectations rather than the expectation of the per-sequence ratio.
Since the window following one that starts at $s$ begins at $s+L_s+1$, and ignoring early
termination and sequence-boundary effects, the start probabilities obey
\begin{equation}
\rho_1=1,
\qquad
\rho_t=\sum_{k=0}^{\gamma}\rho_{t-k-1}\,\mathbb{P}(L_{t-k-1}=k),
\quad t>1,
\label{eq:start_rec}
\end{equation}
with $\rho_{t'}=0$ for $t'<1$. We recompute $\bar{\rho}_t$ from the current draft model at
each training step and stop gradients through the weights.

\subsubsection{Weighted Chain WTV}
\label{sec:weighted_wtv}

For chain drafting under sampling verification,
$a_t=1-\mathrm{TV}(\tilde{p}_t,\tilde{q}_t)$
as in Eq.~\ref{eq:rejection_rate}.
The accepted-length distribution of a window
starting at position $t$ is given by
\begin{equation}
\mathbb{P}(L_t=k)=
\begin{cases}
\displaystyle
\left(\prod_{j=0}^{k-1}a_{t+j}\right)
(1-a_{t+k}), & 0\le k<\gamma,\\[6pt]
\displaystyle
\prod_{j=0}^{\gamma-1}a_{t+j}, & k=\gamma.
\end{cases}
\label{eq:chain_len_dist}
\end{equation}

This distribution determines the window-start weights
$\bar{\rho}_t$ through Eq.~\ref{eq:start_rec}.
Substituting Eq.~\ref{eq:eal_tv} into
Eq.~\ref{eq:weighted_identity} gives the
Weighted Chain WTV objective:
\begin{equation}
\mathcal{L}_{\mathrm{W\text{-}WTV}}(\theta)
=
-\sum_{t=1}^{N}\bar{\rho}_t
\sum_{k=1}^{\gamma}
\prod_{j=0}^{k-1}a_{t+j}.
\label{eq:loss_wwtv}
\end{equation}

Thus, positions more likely to serve as actual
window starts receive larger weights.

\subsubsection{Weighted Tree EAL}
\label{sec:weighted_tree}

We extend the window-start weighting approach to
tree-based speculative decoding. Following
Eq.~\ref{eq:tree_actual_length}, the tree accepted
length is determined by the longest accepted path.
Under greedy verification, the probability of
accepting at least $d$ tokens is
\begin{equation}
\begin{aligned}
&\mathbb{P}(L_{\mathrm{Tree}}^t\ge d)\\
&=
\max_{1\le k\le K}
\left(
\mathbb{I}[d\le L_t(P_k)]
\prod_{j=1}^{d}a_j(k)
\right),
\end{aligned}
\label{eq:tree_accept_survival}
\end{equation}
for $1\le d\le\gamma$, where $\mathbb{I}[\cdot]$
denotes the indicator function.

With $\mathbb{P}(L_{\mathrm{Tree}}^t\ge 0)=1$
and $\mathbb{P}(L_{\mathrm{Tree}}^t\ge\gamma+1)=0$,
the accepted-length distribution is given by
\begin{equation}
\begin{aligned}
&\mathbb{P}(L_{\mathrm{Tree}}^t=d)\\
&=
\mathbb{P}(L_{\mathrm{Tree}}^t\ge d)
-
\mathbb{P}(L_{\mathrm{Tree}}^t\ge d+1),
\end{aligned}
\label{eq:tree_length_dist}
\end{equation}
for $0\le d\le\gamma$.

Substituting this distribution into
Eq.~\ref{eq:start_rec}, with $L_t$ replaced by
$L_{\mathrm{Tree}}^t$, gives the window-start
probabilities $\rho_t$. We then obtain the
normalized weights $\bar{\rho}_t$ following
Eq.~\ref{eq:weighted_identity}.

\begin{equation}
\begin{aligned}
&\mathcal{L}_{\mathrm{W\text{-}TreeEAL}}(\theta)\\
&=
-\sum_{t=1}^{N}
\bar{\rho}_t
\max_{1\le k\le K}
\left(
\sum_{d=1}^{L_t(P_k)}
\prod_{j=1}^{d}a_j(k)
\right).
\end{aligned}
\label{eq:loss_wtree}
\end{equation}

In practice, we approximate the binary conditional
acceptance probabilities $a_j(k)$ using
$q_j^{P_k}(g_j^*)$ to construct differentiable EAL
surrogates and estimate the window-start weights.
The weights are recomputed using the current draft
model and treated as constants during backpropagation,
together with the selected paths and accepted lengths.

\subsubsection{Temporal Tree EAL}
\label{sec:temporal_tree}

Unlike Weighted Tree EAL, which estimates window-start
probabilities, Temporal Tree EAL directly optimizes
the actual window-start positions observed during
greedy tree speculative decoding.

Let $\mathcal{W}_{\mathrm{act}}\subseteq\{1,\ldots,N\}$
denote the recorded window-start positions, with the
first window starting at position $1$.
For each window starting at position $t$, the actual
tree accepted length is $L_{\mathrm{Tree}}^t$, as
defined in Eq.~\ref{eq:tree_actual_length}.
Since the accepted length excludes the additional
target token, the next window starts at position
$t+L_{\mathrm{Tree}}^t+1$, unless decoding has
terminated.

Following Eq.~\ref{eq:tree_expected_length},
we maximize the Tree EAL surrogate over the
recorded window-start positions. The corresponding
Temporal Tree EAL loss is

\begin{equation}
\begin{aligned}
&\mathcal{L}_{\mathrm{TemporalTreeEAL}}(\theta)\\
&=
-\sum_{t\in\mathcal{W}_{\mathrm{act}}}
\max_{1\le k\le K}
\left(
\sum_{d=1}^{L_t(P_k)}
\prod_{j=1}^{d}a_j(k)
\right).
\end{aligned}
\label{eq:temporal_tree_loss}
\end{equation}

Unlike Weighted Tree EAL, which uses probabilistic
window-start weights $\bar{\rho}_t$, Temporal Tree
EAL directly optimizes the recorded window starts
without window-count normalization.

During backpropagation, the recorded window starts,
selected paths, and accepted lengths are treated as
constants. Gradients flow only through the draft
probabilities.

\subsection{GRPO Loss on Acceptance Length}
\label{sec:grpo}

We now describe how we incorporate Group-Relative Policy Optimization
(GRPO)~\cite{shao2024deepseekmath, hu2026bridging} into draft model training, using the speculative decoding
acceptance length as the reward signal. 

\subsubsection{Reward construction via simulated greedy verification}
For a sequence of length $N$, we perform three forward passes: (i) the frozen target model produces the per-position tokens $\{x^{\star}_{j}\}_{j=1}^{N}$ of Eq.~\ref{eq:acc_greedy};
(ii) the current draft model $\pi_{\theta}$ produces per-position $\arg\max$ tokens; (iii) a frozen reference draft $\pi_{\mathrm{ref}}$ (a snapshot of $\pi_{\theta}$ from an earlier training step) produces per-position $\arg\max$ tokens for reward calculation. All three forwards are run once per sequence; individual window-level samples are then obtained by slicing.

For a window of size $\gamma$ starting at position $t$, let
$z^{(t)}_{\theta}, z^{(t)}_{\mathrm{ref}} \in \mathbb{R}^{\gamma \times |\mathcal{V}|}$
be the draft logits of the two policies on positions $t, \ldots, t+\gamma-1$.
The greedy draft trajectory under each policy is
\begin{align}
    S^{\theta,(t)}_{k}        &= \arg\max_{v \in \mathcal{V}} z^{(t)}_{\theta,\,k}(v), \\
    S^{\mathrm{ref},(t)}_{k}  &= \arg\max_{v \in \mathcal{V}} z^{(t)}_{\mathrm{ref},\,k}(v),
    \qquad k = 1, \ldots, \gamma.
\end{align}
Simulating greedy verification against the target, the accepted length
of policy $\pi$ on this window is
\begin{equation}
    \tau^{\pi,(t)}
        \;=\; \sum_{k=1}^{\gamma} \prod_{j=1}^{k} \mathbbm{1}\!\left[S^{\pi,(t)}_{j} = x^{\star}_{t+j-1}\right],
\end{equation}
which counts the number of consecutive correct predictions from the
start of the window until the first mismatch.

\subsubsection{Debiased window reward}
Following GTO~\cite{hu2026bridgingdraftpolicymisalignment}, we debias the reward by
subtracting the reference policy's acceptance length on the same
window:
\begin{equation}
\label{eq:debiased_reward}
    R^{(t)}
    \;=\; \tau^{\pi,(t)} \;-\; \tau^{\mathrm{ref},(t)}.
\end{equation}
Since both policies see the same prefix and are verified against the
same target $\arg\max$ sequence, any prefix-dependent difficulty
cancels out. The reward $R^{(t)}$ therefore captures only the
\emph{incremental} acceptance length that $\pi_{\theta}$ achieves over
the reference at this window. We stop gradients on $R^{(t)}$: gradients
flow only through the policy-ratio term in the surrogate loss, not
through the reward itself.

\subsubsection{Policy ratio along the current trajectory}
For the RL objective we also need the log-probability ratio of
$\pi_{\theta}$ to $\pi_{\mathrm{ref}}$ along the trajectory
$S^{\theta,(t)}$ actually produced by the current policy. We use a
per-token geometric mean to keep the ratio numerically stable across
variable acceptance lengths:
\begin{equation}
    \log \rho^{(t)}
    \;=\; \frac{1}{\gamma} \sum_{k=1}^{\gamma}
        \Big[{\log \pi_{\theta}\!\big(S^{\theta,(t)}_{k} \,\big|\, \cdot\big)
           - \log \pi_{\mathrm{ref}}\!\big(S^{\theta,(t)}_{k} \,\big|\, \cdot\big)}\Big],
\end{equation}
where the conditioning $\cdot$ is the ground-truth prefix
$x_{1:t+k-1}$. We clip $\log \rho^{(t)}$ to $[-10, 10]$ before
exponentiation to avoid overflow.

\subsubsection{Group selection and advantage normalization}
To apply GRPO, window-level samples must be partitioned into
\emph{groups} within which rewards are standardized. For each input
sequence we sample $G$ non-overlapping group anchors uniformly from the valid range. Each group consists of $m$ windows of size $\gamma$ sliced from the same forward pass.
This yields $G \cdot m$ window samples per sequence.

Within each group $g$, let $\{R^{(t)}\}_{t \in g}$ denote the $m$
debiased rewards. We standardize them to form advantages:
\begin{equation}
\label{eq:advantage}
    A^{(t)}
    \;=\; \operatorname{clip}\!\left(
        \frac{R^{(t)} - \bar{R}_{g}}{\operatorname{std}(R_{g}) + \varepsilon},\;
        -C_{A},\; C_{A}
    \right),
    \quad \text{(stop gradient)}
\end{equation}


where $\bar{R}_{g}$ and $\operatorname{std}(R_{g})$ are the group mean
and standard deviation, $\varepsilon = 10^{-6}$ avoids division by
zero, and $C_{A}$ caps extreme advantages.

\subsubsection{GRPO loss}
With advantages in hand, we apply a PPO-style clipped surrogate within
each group. Let $\rho^{(t)} = \exp(\log \rho^{(t)})$ denote the policy
ratio, further clipped to $[\rho_{\min}, \rho_{\max}]$ for
safety. The group loss is
\begin{equation}
    \mathcal{L}^{(g)}_{\mathrm{GRPO}}
    \;=\; -\,\frac{1}{m} \sum_{t \in g}
        \min\!\Big(
            \rho^{(t)} A^{(t)},\;\;
            \operatorname{clip}\!\big(\rho^{(t)},\, 1-\varepsilon_{\mathrm{clip}},\, 1+\varepsilon_{\mathrm{clip}}\big) \, A^{(t)}
        \Big),
\end{equation}
with PPO clipping parameter $\varepsilon_{\mathrm{clip}}$. The
sequence-level GRPO loss averages over the $G$ groups:
\begin{equation}
    \mathcal{L}_{\mathrm{GRPO}}(\theta)
    \;=\; \frac{1}{G} \sum_{g=1}^{G} \mathcal{L}^{(g)}_{\mathrm{GRPO}}.
\end{equation}

\section{Experiments}

\subsection{Settings}

We evaluate our loss designs in a speculative decoding setup with Qwen3~\cite{yang2025qwen3} and Gemma4~\cite{team2026gemma} model families. 
We compare the different losses, including $\mathcal{L}_{\mathrm{EAL}}$ in Section~\ref{sec:eal} and $\mathcal{L}_{\mathrm{WTV}}$ in Section~\ref{sec:wtv} against several baselines and 
configurations. The \textbf{$L_{KL}$} baseline uses only the anchor loss 
$\mathcal{L} = \mathcal{L}_{\mathrm{anchor}}$, where 
$\mathcal{L}_{\mathrm{anchor}}$ is the forward KL divergence between 
the draft and target distributions at each token position. The \textbf{$L_{KL} + L_{TV}$} baseline adds the per-position TV 
distance averaged across the window, 
$\mathcal{L}_{\mathrm{aux}} = \frac{1}{\gamma} \sum_{j=0}^{\gamma-1} \mathrm{TV}(\tilde{p}_{t+j}, \tilde{q}_{t+j})$. The \textbf{$L_{KL} + 0.1L_{EAL}$} baseline adds the differentiable 
expected-acceptance-length surrogate from 
Eq.~\ref{eq:loss_eal} as the auxiliary loss. The \textbf{$L_{LK}$} baseline is the
acceptance-oriented loss of~\citet{samarin2026lk}, which optimizes the per-token
acceptance rate instead of the distributional discrepancy. In contrast, the
\textbf{$L_{WTV}$} loss is trained \emph{without} the anchor loss, using only the proposed
loss as the full training objective. The temperature $T$ of both $\tilde{p}_t$ and
$\tilde{q}_t$ is set to the value used in evaluation.
Rows labeled ``$3$ SFT + $3$ GRPO'' use a two-phase schedule: $3$ epochs of SFT
with the listed supervised objective, followed by $3$ epochs in which the GRPO loss of
Section~\ref{sec:grpo} is added on top with weight $\lambda_{\mathrm{GRPO}} = 0.1$. The
reference policy $\pi_{\mathrm{ref}}$ is the checkpoint at the end of the SFT phase. We
apply this schedule on top of both $L_{KL} + 0.1L_{EAL}$ and $L_{WTV}$, so that the RL
stage is evaluated under both acceptance criteria.

The training dataset consists of 60,000 target-generated sequences: prompts are sampled from ShareGPT, and responses are generated by the frozen target model using greedy decoding. Each training sample concatenates the prompt and the target-generated response and is truncated to a maximum length of 2,048 tokens. All training runs are conducted on two NVIDIA H100 GPUs. We use AdamW with a cosine learning-rate schedule and gradient clipping. For the $L_{KL}+L_{EAL}$ baseline, the peak learning rate is $10^{-5}$. We tune the training hyperparameters separately for each baseline to ensure a fair comparison.

We evaluate on six benchmarks covering different task types: MT-Bench~\cite{zheng2023judging}, GSM8K~\cite{cobbe2021gsm8k}, HumanEval~\cite{chen2021codex}, LongWriter~\cite{bai2025longwriter}, STEM QA~\cite{han2026stemqa}, and SPEED-Bench~\cite{abramovich2026speedbench}.
Specifically, we use a tree of maximum 
depth $\gamma = 7$ with per-depth top-$K$ branching factors 
$(k_1, k_2, k_3, k_4, k_5, k_6, k_7) = (4, 3, 2, 1, 1, 1, 1)$. The fully-expanded tree is then dynamically pruned 
at runtime to a total token budget of $128$ nodes by retaining the 
nodes with the highest cumulative draft scores along their 
root-to-node paths. We also evaluate the chain speculative decoding with $\gamma = 7$. 
Evaluations are conducted on one H100 GPU with greedy and rejection sampling~\cite{li2026breaking} verification, batch size 1, and identical sampling configurations across loss variants to ensure fair comparison. For each benchmark we report acceptance length: the average 
number of tokens accepted per verification step.

We evaluate five target--draft model pairs: Qwen3-8B/Qwen3-0.6B, Qwen3-8B/Qwen3-1.7B, Qwen3-32B/Qwen3-4B, Gemma4-31B/Gemma4-E2B, and gpt-oss-120b/Qwen3-4B~\cite{openai2025gptoss}.

\begin{table}[!h]
\centering
\caption{Acceptance length on six benchmarks with Qwen3-8B as the target model and Qwen3-0.6B as the draft model with tree-based speculative decoding under greedy verification and sampling temperatures $T = 0.5$ and $T = 1.0$. Colored numbers indicate ranking by Avg.\ Acceptance Length within each setting: \textcolor{red}{red} denotes best, \textcolor{blue}{blue} denotes second-best, \textcolor{orange}{orange} denotes third-best.}
\label{tab:main_results}
\renewcommand{\arraystretch}{1.15}
\resizebox{\textwidth}{!}{%
\begin{tabular}{lllcccccccc}
\toprule
\textbf{Setting} & Qwen3-8B & Qwen3-0.6B & \multicolumn{7}{c}{\textbf{Acceptance Length}} & \textbf{Stat.} \\
\cmidrule(lr){2-3} \cmidrule(lr){4-10} \cmidrule(lr){11-11}
 & \textbf{Stage} & \textbf{Loss} & \textbf{MT-Bench} & \textbf{GSM8K} & \textbf{HumanEval} & \textbf{LongWriter} & \textbf{STEM QA} & \textbf{SpeedBench} & \textbf{Avg} & \textbf{Med./P25/P75} \\
\midrule
\multirow{8}{*}{\rotatebox[origin=c]{90}{Greedy}}
 & 6 SFT & $L_{KL}$ & 4.204 & 5.289 & 4.896 & 3.788 & 4.295 &  & 4.494 (0.006) & 5/3/7 \\
 & 6 SFT & \textcolor{orange}{$L_{KL} + 0.1L_{EAL}$} & 4.217 & 5.299 & 4.904 & 3.806 & 4.290 &  & \textcolor{orange}{4.503} (0.007) & 5/3/7 \\
 & 3 SFT + 3 GRPO & \textcolor{red}{$L_{KL} + 0.1L_{EAL} \to L_{KL} + 0.1L_{EAL} + 0.1L_{GRPO}$} & 4.213 & 5.322 & 4.917 & 3.817 & 4.303 &  & \textcolor{red}{4.514} (0.005) & 5/3/7 \\
 & 6 SFT & $L_{CE}$ & 4.180 & 5.309 & 4.885 & 3.749 & 4.286 &  & 4.482 (0.003) & 5/3/7 \\
 & 6 SFT & \textcolor{blue}{$L_{LK}$} & 4.213 & 5.281 & 4.908 & 3.805 & 4.324 &  & \textcolor{blue}{4.506} (0.009) & 5/3/7 \\
 & 6 SFT & $L_{KL} + 0.1L_{TV}$ & 4.198 & 5.297 & 4.898 & 3.802 & 4.291 &  & 4.497 (0.006) & 5/3/7 \\
 & 6 SFT & $L_{WTV}$ & 4.148 & 5.263 & 4.797 & 3.688 & 4.243 &  & 4.428 (0.002) & 5/3/7 \\
 & 6 SFT & $L_{W-WTV}$ &  &  &  &  &  &  &  &  \\
 & 3 SFT + 3 GRPO & $L_{WTV} \to L_{WTV} + 0.1L_{GRPO}$ & 4.169 & 5.341 & 4.878 & 3.724 & 4.267 &  & 4.476 (0.001) & 5/3/7 \\
\midrule
\multirow{6}{*}{\rotatebox[origin=c]{90}{\textbf{$T = 0.5$}}}
 & 6 SFT & $L_{KL}$ & 4.011 & 5.284 & 4.510 & 3.700 & 4.014 &  & 4.304 (0.009) & 5/3/7 \\
 & 6 SFT & $L_{KL} + 0.1L_{EAL}$ & 4.021 & 5.290 & 4.536 & 3.719 & 3.998 &  & 4.313 (0.005) & 5/3/7 \\
 & 3 SFT + 3 GRPO & \textcolor{orange}{$L_{KL} + 0.1L_{EAL} \to L_{KL} + 0.1L_{EAL} + 0.1L_{GRPO}$} & 4.044 & 5.295 & 4.551 & 3.730 & 4.009 &  & \textcolor{orange}{4.326} (0.005) & 5/3/7 \\
 & 6 SFT & $L_{KL} + 0.1L_{TV}$ & 4.010 & 5.278 & 4.522 & 3.707 & 4.004 &  & 4.304 (0.006) & 5/3/7 \\
 & 6 SFT & \textcolor{blue}{$L_{WTV}$} & 4.472 & 5.671 & 5.012 & 4.231 & 4.512 &  & \textcolor{blue}{4.780} (0.018) & 6/3/7 \\
 & 6 SFT & $L_{W-WTV}$ &  &  &  &  &  &  &  &  \\
 & 3 SFT + 3 GRPO & \textcolor{red}{$L_{WTV} \to L_{WTV} + 0.1L_{GRPO}$} & 4.506 & 5.722 & 5.005 & 4.244 & 4.504 &  & \textcolor{red}{4.796} (0.010) & 6/3/7 \\
\midrule
\multirow{8}{*}{\rotatebox[origin=c]{90}{\textbf{$T = 1.0$}}}
 & 6 SFT & $L_{KL}$ & 4.805 & 5.794 & 4.913 & 4.765 & 4.712 &  & 4.998 (0.018) & 6/3/7 \\
 & 6 SFT & $L_{KL} + 0.1L_{EAL}$ & 4.829 & 5.750 & 4.951 & 4.825 & 4.770 &  & 5.025 (0.011) & 6/3/7 \\
 & 3 SFT + 3 GRPO & $L_{KL} + 0.1L_{EAL} \to L_{KL} + 0.1L_{EAL} + 0.1L_{GRPO}$ & 4.802 & 5.849 & 4.903 & 4.829 & 4.778 &  & 5.032 (0.003) & 6/3/7 \\
 & 6 SFT & \textcolor{orange}{$L_{CE}$} & 4.805 & 5.882 & 5.038 & 4.795 & 4.816 &  & \textcolor{orange}{5.067} (0.010) & 6/4/7 \\
 & 6 SFT & $L_{LK}$ & 4.808 & 5.798 & 4.918 & 4.731 & 4.713 &  & 4.994 (0.012) & 6/3/7 \\
 & 6 SFT & $L_{KL} + 0.1L_{TV}$ & 4.742 & 5.800 & 4.940 & 4.748 & 4.621 &  & 4.970 (0.004) & 6/4/7 \\
 & 6 SFT & \textcolor{blue}{$L_{WTV}$} & 5.580 & 6.436 & 5.687 & 5.629 & 5.416 &  & \textcolor{blue}{5.750} (0.012) & 7/5/7 \\
 & 6 SFT & $L_{W-WTV}$ &  &  &  &  &  &  &  &  \\
 & 3 SFT + 3 GRPO & \textcolor{red}{$L_{WTV} \to L_{WTV} + 0.1L_{GRPO}$} & 5.596 & 6.453 & 5.697 & 5.599 & 5.472 &  & \textcolor{red}{5.763} (0.019) & 7/5/7 \\
\bottomrule
\end{tabular}%
}
\end{table}

\begin{table}[!h]
\centering
\caption{Acceptance length on six benchmarks with Qwen3-8B as the target model and Qwen3-0.6B as the draft model with chain-based speculative decoding under greedy verification and sampling temperatures $T = 0.5$ and $T = 1.0$. Colored numbers indicate ranking by Avg.\ Acceptance Length within each setting: \textcolor{red}{red} denotes best, \textcolor{blue}{blue} denotes second-best, \textcolor{orange}{orange} denotes third-best.}
\label{tab:main_results_chain}
\renewcommand{\arraystretch}{1.15}
\resizebox{\textwidth}{!}{%
\begin{tabular}{lllcccccccc}
\toprule
\textbf{Setting} & Qwen3-8B & Qwen3-0.6B & \multicolumn{7}{c}{\textbf{Acceptance Length}} & \textbf{Stat.} \\
\cmidrule(lr){2-3} \cmidrule(lr){4-10} \cmidrule(lr){11-11}
 & \textbf{Stage} & \textbf{Loss} & \textbf{MT-Bench} & \textbf{GSM8K} & \textbf{HumanEval} & \textbf{LongWriter} & \textbf{STEM QA} & \textbf{SpeedBench} & \textbf{Avg} & \textbf{Med./P25/P75} \\
\midrule
\multirow{8}{*}{\rotatebox[origin=c]{90}{Greedy}}
 & 6 SFT & $L_{KL}$ & 2.670 & 3.826 & 3.268 & 2.573 & 2.919 &  & 3.051 (0.002) & 3/1/7 \\
 & 6 SFT & \textcolor{blue}{$L_{KL} + 0.1L_{EAL}$} & 2.682 & 3.837 & 3.272 & 2.618 & 2.912 &  & \textcolor{blue}{3.064} (0.005) & 3/1/7 \\
 & 3 SFT + 3 GRPO & $L_{KL} + 0.1L_{EAL} \to L_{KL} + 0.1L_{EAL} + 0.1L_{GRPO}$ & 2.679 & 3.849 & 3.279 & 2.597 & 2.900 &  & 3.061 (0.012) & 3/1/7 \\
 & 6 SFT & $L_{CE}$ & 2.685 & 3.815 & 3.228 & 2.569 & 2.926 &  & 3.045 (0.003) & 2/1/6 \\
 & 6 SFT & $L_{LK}$ & 2.675 & 3.831 & 3.270 & 2.594 & 2.922 &  & 3.058 (0.004) & 3/1/7 \\
 & 6 SFT & $L_{KL} + 0.1L_{TV}$ & 2.676 & 3.833 & 3.273 & 2.581 & 2.945 &  & 3.062 (0.004) & 3/1/7 \\
 & 6 SFT & \textcolor{orange}{$L_{WTV}$} & 2.712 & 3.856 & 3.215 & 2.571 & 2.965 &  & \textcolor{orange}{3.064} (0.003) & 3/1/7 \\
 & 6 SFT & $L_{W-WTV}$ &  &  &  &  &  &  &  &  \\
 & 3 SFT + 3 GRPO & \textcolor{red}{$L_{WTV} \to L_{WTV} + 0.1L_{GRPO}$} & 2.715 & 3.861 & 3.220 & 2.578 & 2.952 &  & \textcolor{red}{3.065} (0.008) & 3/1/7 \\
\midrule
\multirow{7}{*}{\rotatebox[origin=c]{90}{\textbf{$T = 0.5$}}}
 & 6 SFT & $L_{KL}$ & 2.614 & 3.811 & 2.943 & 2.515 & 2.726 &  & 2.922 (0.011) & 2/1/6 \\
 & 6 SFT & $L_{KL} + 0.1L_{EAL}$ & 2.645 & 3.799 & 2.978 & 2.515 & 2.732 &  & 2.934 (0.014) & 2/1/6 \\
 & 3 SFT + 3 GRPO & $L_{KL} + 0.1L_{EAL} \to L_{KL} + 0.1L_{EAL} + 0.1L_{GRPO}$ & 2.656 & 3.844 & 3.031 & 2.545 & 2.786 &  & 2.972 (0.022) & 2/1/6 \\
 & 6 SFT & \textcolor{orange}{$L_{CE}$} & 2.651 & 3.846 & 3.042 & 2.574 & 2.776 &  & \textcolor{orange}{2.978} (0.019) & 2/1/6 \\
 & 6 SFT & $L_{KL} + 0.1L_{TV}$ & 2.606 & 3.779 & 3.009 & 2.512 & 2.736 &  & 2.928 (0.009) & 2/1/6 \\
 & 6 SFT & \textcolor{blue}{$L_{WTV}$} & 2.712 & 3.929 & 3.218 & 2.636 & 2.853 &  & \textcolor{blue}{3.070} (0.024) & 3/1/7 \\
 & 6 SFT & $L_{W-WTV}$ &  &  &  &  &  &  &  &  \\
 & 3 SFT + 3 GRPO & \textcolor{red}{$L_{WTV} \to L_{WTV} + 0.1L_{GRPO}$} & 2.770 & 3.883 & 3.221 & 2.651 & 2.899 &  & \textcolor{red}{3.085} (0.021) & 3/1/7 \\
\midrule
\multirow{8}{*}{\rotatebox[origin=c]{90}{\textbf{$T = 1.0$}}}
 & 6 SFT & $L_{KL}$ & 2.407 & 3.698 & 2.658 & 2.689 & 2.665 &  & 2.823 (0.011) & 2/0/5 \\
 & 6 SFT & $L_{KL} + 0.1L_{EAL}$ & 2.459 & 3.692 & 2.681 & 2.710 & 2.678 &  & 2.844 (0.024) & 2/0/5 \\
 & 3 SFT + 3 GRPO & $L_{KL} + 0.1L_{EAL} \to L_{KL} + 0.1L_{EAL} + 0.1L_{GRPO}$ & 2.424 & 3.797 & 2.732 & 2.755 & 2.715 &  & 2.885 (0.014) & 2/0/5 \\
 & 6 SFT & \textcolor{orange}{$L_{CE}$} & 2.472 & 3.768 & 2.745 & 2.739 & 2.707 &  & \textcolor{orange}{2.886} (0.021) & 2/0/5 \\
 & 6 SFT & $L_{LK}$ & 2.431 & 3.695 & 2.667 & 2.681 & 2.673 &  & 2.829 (0.016) & 2/0/5 \\
 & 6 SFT & $L_{KL} + 0.1L_{TV}$ & 2.406 & 3.660 & 2.702 & 2.712 & 2.675 &  & 2.831 (0.015) & 2/0/5 \\
 & 6 SFT & \textcolor{blue}{$L_{WTV}$} & 2.692 & 3.960 & 3.190 & 2.986 & 3.017 &  & \textcolor{blue}{3.169} (0.020) & 3/1/7 \\
 & 6 SFT & $L_{W-WTV}$ &  &  &  &  &  &  &  &  \\
 & 3 SFT + 3 GRPO & \textcolor{red}{$L_{WTV} \to L_{WTV} + 0.1L_{GRPO}$} & 2.707 & 3.989 & 3.223 & 3.055 & 3.043 &  & \textcolor{red}{3.203} (0.021) & 3/1/7 \\
\bottomrule
\end{tabular}%
}
\end{table}

\subsection{Results}


Table~\ref{tab:main_results} reports the acceptance length for Qwen3-8B/Qwen3-0.6B under
tree drafting. Under greedy verification the acceptance-oriented objectives cluster
tightly: $L_{KL}+0.1L_{EAL}$ reaches $4.503$ against $4.494$ for $L_{KL}$, the subsequent
GRPO stage gives the best result at $4.514$, and $L_{LK}$ is competitive at $4.506$.
$L_{WTV}$ is the weakest objective here at $4.428$, consistent with the fact that its
acceptance model does not match greedy verification. The picture reverses under sampling
verification. At $T=0.5$, $L_{WTV}$ reaches $4.780$ against $4.304$ for $L_{KL}$ and
$4.313$ for $L_{KL}+0.1L_{EAL}$, and at $T=1.0$ it reaches $5.750$ against $4.998$; adding
GRPO on top of WTV improves these to $4.796$ and $5.763$. 

Table~\ref{tab:main_results_chain} reports the same pair under chain drafting. Under greedy
verification the spread between objectives is small: $L_{KL}+0.1L_{EAL}$ and $L_{WTV}$ both
reach $3.064$ against $3.051$ for $L_{KL}$, and the GRPO stage on top of WTV gives the best
result at $3.065$. Under sampling verification the gap widens: $L_{WTV}$
reaches $3.070$ at $T=0.5$ and $3.169$ at $T=1.0$, against $2.922$ and $2.823$ for
$L_{KL}$, with GRPO adding a further gain to $3.085$ and $3.203$.

\begin{table}[!h]
\centering
\caption{Acceptance length on six benchmarks with Qwen3-8B as the target model and Qwen3-1.7B as the draft model with tree-based speculative decoding under greedy verification and sampling temperatures $T = 0.5$ and $T = 1.0$. Colored numbers indicate ranking by Avg.\ Acceptance Length within each setting: \textcolor{red}{red} denotes best, \textcolor{blue}{blue} denotes second-best, \textcolor{orange}{orange} denotes third-best.}
\label{tab:sd2_merged}
\renewcommand{\arraystretch}{1.15}
\resizebox{\textwidth}{!}{%
\begin{tabular}{lllcccccccc}
\toprule
\textbf{Setting} & Qwen3-8B & Qwen3-1.7B & \multicolumn{7}{c}{\textbf{Acceptance Length}} & \textbf{Stat.} \\
\cmidrule(lr){2-3} \cmidrule(lr){4-10} \cmidrule(lr){11-11}
 & \textbf{Stage} & \textbf{Loss} & \textbf{MT-Bench} & \textbf{GSM8K} & \textbf{HumanEval} & \textbf{LongWriter} & \textbf{STEM QA} & \textbf{SpeedBench} & \textbf{Avg} & \textbf{Med./P25/P75} \\
\midrule
\multirow{8}{*}{\rotatebox[origin=c]{90}{Greedy}}
 & 6 SFT & \textcolor{orange}{$L_{KL}$} & 4.840 & 5.705 & 5.372 & 4.407 & 4.915 &  & \textcolor{orange}{5.048} (0.001) & 6/4/7 \\
 & 6 SFT & \textcolor{blue}{$L_{KL} + 0.1L_{EAL}$} & 4.872 & 5.700 & 5.375 & 4.425 & 4.914 &  & \textcolor{blue}{5.057} (0.004) & 6/4/7 \\
 & 3 SFT + 3 GRPO & \textcolor{red}{$L_{KL} + 0.1L_{EAL} \to L_{KL} + 0.1L_{EAL} + 0.1L_{GRPO}$} & 4.888 & 5.741 & 5.410 & 4.391 & 4.978 &  & \textcolor{red}{5.082} (0.009) & 7/4/7 \\
 & 6 SFT & $L_{CE}$ & 4.803 & 5.648 & 5.359 & 4.335 & 4.882 &  & 5.005 (0.004) & 6/3/7 \\
 & 6 SFT & $L_{LK}$ & 4.852 & 5.703 & 5.373 & 4.341 & 4.919 &  & 5.038 (0.006) & 6/4/7 \\
 & 6 SFT & $L_{KL} + 0.1L_{TV}$ & 4.834 & 5.726 & 5.372 & 4.341 & 4.933 &  & 5.041 (0.008) & 6/4/7 \\
 & 6 SFT & $L_{WTV}$ & 4.803 & 5.726 & 5.379 & 4.350 & 4.957 &  & 5.043 (0.005) & 6/4/7 \\
 & 6 SFT & $L_{W-WTV}$ &  &  &  &  &  &  &  &  \\
 & 3 SFT + 3 GRPO & $L_{WTV} \to L_{WTV} + 0.1L_{GRPO}$ & 4.828 & 5.727 & 5.385 & 4.340 & 4.920 &  & 5.040 (0.016) & 6/4/7 \\
\midrule
\multirow{6}{*}{\rotatebox[origin=c]{90}{\textbf{$T = 0.5$}}}
 & 6 SFT & $L_{KL}$ & 4.793 & 5.728 & 5.225 & 4.552 & 4.723 &  & 5.004 (0.010) & 6/4/7 \\
 & 6 SFT & $L_{KL} + 0.1L_{EAL}$ & 4.793 & 5.746 & 5.256 & 4.538 & 4.710 &  & 5.009 (0.009) & 6/4/7 \\
 & 3 SFT + 3 GRPO & \textcolor{orange}{$L_{KL} + 0.1L_{EAL} \to L_{KL} + 0.1L_{EAL} + 0.1L_{GRPO}$} & 4.856 & 5.729 & 5.254 & 4.552 & 4.732 &  & \textcolor{orange}{5.025} (0.017) & 6/4/7 \\
 & 6 SFT & $L_{KL} + 0.1L_{TV}$ & 4.818 & 5.727 & 5.221 & 4.533 & 4.700 &  & 5.000 (0.008) & 6/4/7 \\
 & 6 SFT & \textcolor{blue}{$L_{WTV}$} & 5.200 & 6.055 & 5.602 & 4.999 & 5.240 &  & \textcolor{blue}{5.419} (0.012) & 7/4/7 \\
 & 6 SFT & $L_{W-WTV}$ &  &  &  &  &  &  &  &  \\
 & 3 SFT + 3 GRPO & \textcolor{red}{$L_{WTV} \to L_{WTV} + 0.1L_{GRPO}$} & 5.218 & 6.069 & 5.634 & 5.005 & 5.233 &  & \textcolor{red}{5.432} (0.007) & 7/5/7 \\
\midrule
\multirow{8}{*}{\rotatebox[origin=c]{90}{\textbf{$T = 1.0$}}}
 & 6 SFT & $L_{KL}$ & 5.644 & 6.265 & 5.673 & 5.678 & 5.385 &  & 5.729 (0.019) & 7/5/7 \\
 & 6 SFT & $L_{KL} + 0.1L_{EAL}$ & 5.678 & 6.309 & 5.670 & 5.635 & 5.359 &  & 5.730 (0.013) & 7/5/7 \\
 & 3 SFT + 3 GRPO & $L_{KL} + 0.1L_{EAL} \to L_{KL} + 0.1L_{EAL} + 0.1L_{GRPO}$ & 5.615 & 6.304 & 5.768 & 5.667 & 5.419 &  & 5.755 (0.016) & 7/5/7 \\
 & 6 SFT & \textcolor{orange}{$L_{CE}$} & 5.982 & 6.539 & 6.213 & 6.049 & 5.804 &  & \textcolor{orange}{6.117} (0.014) & 7/6/7 \\
 & 6 SFT & $L_{LK}$ & 5.662 & 6.284 & 5.671 & 5.696 & 5.504 &  & 5.763 (0.015) & 7/5/7 \\
 & 6 SFT & $L_{KL} + 0.1L_{TV}$ & 5.661 & 6.293 & 5.695 & 5.654 & 5.373 &  & 5.735 (0.009) & 7/5/7 \\
 & 6 SFT & \textcolor{blue}{$L_{WTV}$} & 6.272 & 6.775 & 6.419 & 6.350 & 6.256 &  & \textcolor{blue}{6.414} (0.023) & 7/7/7 \\
 & 6 SFT & $L_{W-WTV}$ &  &  &  &  &  &  &  &  \\
 & 3 SFT + 3 GRPO & \textcolor{red}{$L_{WTV} \to L_{WTV} + 0.1L_{GRPO}$} & 6.379 & 6.764 & 6.382 & 6.413 & 6.345 &  & \textcolor{red}{6.457} (0.018) & 7/7/7 \\
\bottomrule
\end{tabular}%
}
\end{table}

\begin{table}[!h]
\centering
\caption{Acceptance length on six benchmarks with Qwen3-8B as the target model and Qwen3-1.7B as the draft model with chain-based speculative decoding under greedy verification and sampling temperatures $T = 0.5$ and $T = 1.0$. Colored numbers indicate ranking by Avg.\ Acceptance Length within each setting: \textcolor{red}{red} denotes best, \textcolor{blue}{blue} denotes second-best, \textcolor{orange}{orange} denotes third-best.}
\label{tab:sd2_merged_chain}
\renewcommand{\arraystretch}{1.15}
\resizebox{\textwidth}{!}{%
\begin{tabular}{lllcccccccc}
\toprule
\textbf{Setting} & Qwen3-8B & Qwen3-1.7B & \multicolumn{7}{c}{\textbf{Acceptance Length}} & \textbf{Stat.} \\
\cmidrule(lr){2-3} \cmidrule(lr){4-10} \cmidrule(lr){11-11}
 & \textbf{Stage} & \textbf{Loss} & \textbf{MT-Bench} & \textbf{GSM8K} & \textbf{HumanEval} & \textbf{LongWriter} & \textbf{STEM QA} & \textbf{SpeedBench} & \textbf{Avg} & \textbf{Med./P25/P75} \\
\midrule
\multirow{8}{*}{\rotatebox[origin=c]{90}{Greedy}}
 & 6 SFT & $L_{KL}$ & 3.265 & 4.399 & 3.859 & 3.183 & 3.537 &  & 3.649 (0.001) & 4/1/7 \\
 & 6 SFT & \textcolor{orange}{$L_{KL} + 0.1L_{EAL}$} & 3.274 & 4.413 & 3.871 & 3.180 & 3.551 &  & \textcolor{orange}{3.658} (0.004) & 4/1/7 \\
 & 3 SFT + 3 GRPO & \textcolor{red}{$L_{KL} + 0.1L_{EAL} \to L_{KL} + 0.1L_{EAL} + 0.1L_{GRPO}$} & 3.320 & 4.431 & 3.892 & 3.250 & 3.598 &  & \textcolor{red}{3.698} (0.015) & 4/1/7 \\
 & 6 SFT & $L_{CE}$ & 3.108 & 4.295 & 3.794 & 3.074 & 3.458 &  & 3.546 (0.006) & 3/1/7 \\
 & 6 SFT & $L_{LK}$ & 3.270 & 4.405 & 3.864 & 3.202 & 3.535 &  & 3.655 (0.003) & 4/1/7 \\
 & 6 SFT & $L_{KL} + 0.1L_{TV}$ & 3.260 & 4.399 & 3.871 & 3.188 & 3.550 &  & 3.654 (0.004) & 4/1/7 \\
 & 6 SFT & \textcolor{blue}{$L_{WTV}$} & 3.267 & 4.396 & 3.858 & 3.179 & 3.593 &  & \textcolor{blue}{3.659} (0.002) & 4/1/7 \\
 & 6 SFT & $L_{W-WTV}$ &  &  &  &  &  &  &  &  \\
 & 3 SFT + 3 GRPO & $L_{WTV} \to L_{WTV} + 0.1L_{GRPO}$ & 3.295 & 4.393 & 3.869 & 3.155 & 3.564 &  & 3.655 (0.013) & 4/1/7 \\
\midrule
\multirow{7}{*}{\rotatebox[origin=c]{90}{\textbf{$T = 0.5$}}}
 & 6 SFT & $L_{KL}$ & 3.273 & 4.465 & 3.671 & 3.299 & 3.413 &  & 3.624 (0.005) & 4/1/7 \\
 & 6 SFT & $L_{KL} + 0.1L_{EAL}$ & 3.299 & 4.461 & 3.671 & 3.306 & 3.458 &  & 3.639 (0.004) & 4/1/7 \\
 & 3 SFT + 3 GRPO & \textcolor{orange}{$L_{KL} + 0.1L_{EAL} \to L_{KL} + 0.1L_{EAL} + 0.1L_{GRPO}$} & 3.317 & 4.464 & 3.700 & 3.293 & 3.427 &  & \textcolor{orange}{3.640} (0.013) & 4/1/7 \\
 & 6 SFT & $L_{CE}$ & 3.154 & 4.278 & 3.702 & 3.227 & 3.302 &  & 3.533 (0.027) & 3/1/7 \\
 & 6 SFT & $L_{KL} + 0.1L_{TV}$ & 3.294 & 4.388 & 3.726 & 3.261 & 3.447 &  & 3.623 (0.005) & 4/1/7 \\
 & 6 SFT & \textcolor{blue}{$L_{WTV}$} & 3.312 & 4.481 & 3.774 & 3.311 & 3.475 &  & \textcolor{blue}{3.671} (0.018) & 4/1/7 \\
 & 6 SFT & $L_{W-WTV}$ &  &  &  &  &  &  &  &  \\
 & 3 SFT + 3 GRPO & \textcolor{red}{$L_{WTV} \to L_{WTV} + 0.1L_{GRPO}$} & 3.348 & 4.480 & 3.807 & 3.311 & 3.433 &  & \textcolor{red}{3.676} (0.010) & 4/1/7 \\
\midrule
\multirow{8}{*}{\rotatebox[origin=c]{90}{\textbf{$T = 1.0$}}}
 & 6 SFT & $L_{KL}$ & 3.122 & 4.340 & 3.554 & 3.554 & 3.394 &  & 3.593 (0.014) & 3/1/7 \\
 & 6 SFT & $L_{KL} + 0.1L_{EAL}$ & 3.132 & 4.389 & 3.557 & 3.564 & 3.337 &  & 3.596 (0.017) & 3/1/7 \\
 & 3 SFT + 3 GRPO & $L_{KL} + 0.1L_{EAL} \to L_{KL} + 0.1L_{EAL} + 0.1L_{GRPO}$ & 3.140 & 4.389 & 3.583 & 3.568 & 3.407 &  & 3.617 (0.011) & 3/1/7 \\
 & 6 SFT & \textcolor{orange}{$L_{CE}$} & 3.173 & 4.372 & 3.748 & 3.604 & 3.436 &  & \textcolor{orange}{3.667} (0.036) & 3/1/7 \\
 & 6 SFT & $L_{LK}$ & 3.127 & 4.362 & 3.555 & 3.544 & 3.410 &  & 3.600 (0.015) & 3/1/7 \\
 & 6 SFT & $L_{KL} + 0.1L_{TV}$ & 3.125 & 4.352 & 3.500 & 3.529 & 3.359 &  & 3.573 (0.016) & 3/1/7 \\
 & 6 SFT & \textcolor{blue}{$L_{WTV}$} & 3.252 & 4.553 & 3.787 & 3.707 & 3.537 &  & \textcolor{blue}{3.767} (0.029) & 4/1/7 \\
 & 6 SFT & $L_{W-WTV}$ &  &  &  &  &  &  &  &  \\
 & 3 SFT + 3 GRPO & \textcolor{red}{$L_{WTV} \to L_{WTV} + 0.1L_{GRPO}$} & 3.299 & 4.549 & 3.792 & 3.757 & 3.549 &  & \textcolor{red}{3.789} (0.009) & 4/1/7 \\
\bottomrule
\end{tabular}%
}
\end{table}

Table~\ref{tab:sd2_merged} reports the acceptance length for Qwen3-8B/Qwen3-1.7B under tree
drafting. With this larger draft model the greedy column compresses further: $L_{KL}$
itself reaches $5.048$, $L_{KL}+0.1L_{EAL}$ $5.057$, and the GRPO stage $5.082$, while the
remaining objectives all fall within $0.05$ of the baseline. The advantage of WTV again
appears only under sampling verification, where it reaches $5.419$ at $T=0.5$ and $6.414$
at $T=1.0$ against $5.004$ and $5.729$ for $L_{KL}$, with GRPO improving these to $5.432$
and $6.457$.

Table~\ref{tab:sd2_merged_chain} reports the same pair under chain drafting. Under greedy
verification the GRPO stage on top of $L_{KL}+0.1L_{EAL}$ gives the best result at $3.698$,
ahead of $L_{WTV}$ at $3.659$ and $L_{KL}+0.1L_{EAL}$ at $3.658$; $L_{CE}$ is a clear
outlier at $3.546$. Under sampling verification $L_{WTV}$ leads at both temperatures,
reaching $3.671$ at $T=0.5$ and $3.767$ at $T=1.0$ against $3.624$ and $3.593$ for
$L_{KL}$, and GRPO improves these to $3.676$ and $3.789$.

\begin{table}[!h]
\centering
\caption{Acceptance length on six benchmarks with Qwen3-32B as the target model and Qwen3-4B as the draft model with tree-based speculative decoding under greedy verification and sampling temperatures $T = 0.5$ and $T = 1.0$. Colored numbers indicate ranking by Avg.\ Acceptance Length within each setting: \textcolor{red}{red} denotes best, \textcolor{blue}{blue} denotes second-best, \textcolor{orange}{orange} denotes third-best.}
\label{tab:sd3_merged}
\renewcommand{\arraystretch}{1.15}
\resizebox{\textwidth}{!}{%
\begin{tabular}{lllcccccccc}
\toprule
\textbf{Setting} & Qwen3-32B & Qwen3-4B & \multicolumn{7}{c}{\textbf{Acceptance Length}} & \textbf{Stat.} \\
\cmidrule(lr){2-3} \cmidrule(lr){4-10} \cmidrule(lr){11-11}
 & \textbf{Stage} & \textbf{Loss} & \textbf{MT-Bench} & \textbf{GSM8K} & \textbf{HumanEval} & \textbf{LongWriter} & \textbf{STEM QA} & \textbf{SpeedBench} & \textbf{Avg} & \textbf{Med./P25/P75} \\
\midrule
\multirow{7}{*}{\rotatebox[origin=c]{90}{Greedy}}
 & 6 SFT & $L_{KL}$ & 4.911 & 5.827 & 5.081 & 4.758 & 4.709 &  & 5.057 (0.005) & 6/4/7 \\
 & 6 SFT & \textcolor{blue}{$L_{KL} + 0.1L_{EAL}$} & 4.915 & 5.843 & 5.106 & 4.792 & 4.709 &  & \textcolor{blue}{5.073} (0.008) & 6/4/7 \\
 & 3 SFT + 3 GRPO & \textcolor{red}{$L_{KL} + 0.1L_{EAL} \to L_{KL} + 0.1L_{EAL} + 0.1L_{GRPO}$} & 4.956 & 5.839 & 5.108 & 4.805 & 4.732 &  & \textcolor{red}{5.088} (0.013) & 6/4/7 \\
 & 6 SFT & $L_{CE}$ & 4.886 & 5.803 & 5.047 & 4.766 & 4.680 &  & 5.036 (0.007) & 6/3/7 \\
 & 6 SFT & \textcolor{orange}{$L_{LK}$} & 4.913 & 5.834 & 5.092 & 4.744 & 4.732 &  & \textcolor{orange}{5.063} (0.006) & 6/4/7 \\
 & 6 SFT & $L_{KL} + 0.1L_{TV}$ & 4.893 & 5.844 & 5.067 & 4.757 & 4.725 &  & 5.057 (0.007) & 6/3/7 \\
 & 6 SFT & $L_{WTV}$ & 4.901 & 5.796 & 5.066 & 4.742 & 4.690 &  & 5.039 (0.011) & 6/3/7 \\
 & 6 SFT & $L_{W-WTV}$ &  &  &  &  &  &  &  &  \\
\midrule
\multirow{5}{*}{\rotatebox[origin=c]{90}{\textbf{$T = 0.5$}}}
 & 6 SFT & $L_{KL}$ & 4.891 & 5.770 & 4.851 & 4.877 & 4.485 &  & 4.975 (0.007) & 6/3/7 \\
 & 6 SFT & \textcolor{orange}{$L_{KL} + 0.1L_{EAL}$} & 4.914 & 5.731 & 4.896 & 4.893 & 4.495 &  & \textcolor{orange}{4.986} (0.011) & 6/3/7 \\
 & 3 SFT + 3 GRPO & \textcolor{blue}{$L_{KL} + 0.1L_{EAL} \to L_{KL} + 0.1L_{EAL} + 0.1L_{GRPO}$} & 4.981 & 5.742 & 4.918 & 4.892 & 4.500 &  & \textcolor{blue}{5.007} (0.004) & 6/4/7 \\
 & 6 SFT & $L_{KL} + 0.1L_{TV}$ & 4.929 & 5.750 & 4.842 & 4.879 & 4.496 &  & 4.979 (0.006) & 6/3/7 \\
 & 6 SFT & \textcolor{red}{$L_{WTV}$} & 5.394 & 6.150 & 5.582 & 5.296 & 5.057 &  & \textcolor{red}{5.496} (0.028) & 7/4/7 \\
 & 6 SFT & $L_{W-WTV}$ &  &  &  &  &  &  &  &  \\
\midrule
\multirow{7}{*}{\rotatebox[origin=c]{90}{\textbf{$T = 1.0$}}}
 & 6 SFT & $L_{KL}$ & 6.166 & 6.471 & 5.318 & 6.325 & 5.519 &  & 5.960 (0.015) & 7/5/7 \\
 & 6 SFT & $L_{KL} + 0.1L_{EAL}$ & 6.231 & 6.455 & 5.401 & 6.364 & 5.591 &  & 6.008 (0.017) & 7/5/7 \\
 & 3 SFT + 3 GRPO & \textcolor{orange}{$L_{KL} + 0.1L_{EAL} \to L_{KL} + 0.1L_{EAL} + 0.1L_{GRPO}$} & 6.237 & 6.491 & 5.421 & 6.396 & 5.606 &  & \textcolor{orange}{6.030} (0.021) & 7/5/7 \\
 & 6 SFT & \textcolor{blue}{$L_{CE}$} & 6.590 & 6.859 & 6.242 & 6.840 & 6.013 &  & \textcolor{blue}{6.509} (0.037) & 7/7/7 \\
 & 6 SFT & $L_{LK}$ & 6.197 & 6.463 & 5.357 & 6.241 & 5.554 &  & 5.962 (0.016) & 7/5/7 \\
 & 6 SFT & $L_{KL} + 0.1L_{TV}$ & 6.181 & 6.480 & 5.312 & 6.310 & 5.519 &  & 5.960 (0.019) & 7/5/7 \\
 & 6 SFT & \textcolor{red}{$L_{WTV}$} & 6.684 & 6.895 & 6.724 & 6.632 & 6.422 &  & \textcolor{red}{6.671} (0.005) & 7/7/7 \\
 & 6 SFT & $L_{W-WTV}$ &  &  &  &  &  &  &  &  \\
\bottomrule
\end{tabular}%
}
\end{table}

\begin{table}[!h]
\centering
\caption{Acceptance length on six benchmarks with Qwen3-32B as the target model and Qwen3-4B as the draft model with chain-based speculative decoding under greedy verification and sampling temperatures $T = 0.5$ and $T = 1.0$. Colored numbers indicate ranking by Avg.\ Acceptance Length within each setting: \textcolor{red}{red} denotes best, \textcolor{blue}{blue} denotes second-best, \textcolor{orange}{orange} denotes third-best.}
\label{tab:sd3_merged_chain}
\renewcommand{\arraystretch}{1.15}
\resizebox{\textwidth}{!}{%
\begin{tabular}{lllcccccccc}
\toprule
\textbf{Setting} & Qwen3-32B & Qwen3-4B & \multicolumn{7}{c}{\textbf{Acceptance Length}} & \textbf{Stat.} \\
\cmidrule(lr){2-3} \cmidrule(lr){4-10} \cmidrule(lr){11-11}
 & \textbf{Stage} & \textbf{Loss} & \textbf{MT-Bench} & \textbf{GSM8K} & \textbf{HumanEval} & \textbf{LongWriter} & \textbf{STEM QA} & \textbf{SpeedBench} & \textbf{Avg} & \textbf{Med./P25/P75} \\
\midrule
\multirow{7}{*}{\rotatebox[origin=c]{90}{Greedy}}
 & 6 SFT & \textcolor{orange}{$L_{KL}$} & 3.355 & 4.544 & 3.412 & 3.398 & 3.365 &  & \textcolor{orange}{3.615} (0.010) & 3/1/7 \\
 & 6 SFT & \textcolor{blue}{$L_{KL} + 0.1L_{EAL}$} & 3.348 & 4.536 & 3.462 & 3.425 & 3.387 &  & \textcolor{blue}{3.632} (0.005) & 3/1/7 \\
 & 3 SFT + 3 GRPO & \textcolor{red}{$L_{KL} + 0.1L_{EAL} \to L_{KL} + 0.1L_{EAL} + 0.1L_{GRPO}$} & 3.368 & 4.501 & 3.485 & 3.440 & 3.385 &  & \textcolor{red}{3.636} (0.008) & 3/1/7 \\
 & 6 SFT & $L_{CE}$ & 3.238 & 4.412 & 3.369 & 3.310 & 3.280 &  & 3.522 (0.026) & 3/1/7 \\
 & 6 SFT & $L_{LK}$ & 3.351 & 4.540 & 3.428 & 3.367 & 3.386 &  & 3.614 (0.008) & 3/1/7 \\
 & 6 SFT & $L_{KL} + 0.1L_{TV}$ & 3.358 & 4.516 & 3.442 & 3.380 & 3.360 &  & 3.611 (0.012) & 3/1/7 \\
 & 6 SFT & $L_{WTV}$ & 3.375 & 4.522 & 3.425 & 3.356 & 3.337 &  & 3.603 (0.026) & 3/1/7 \\
 & 6 SFT & $L_{W-WTV}$ &  &  &  &  &  &  &  &  \\
\midrule
\multirow{5}{*}{\rotatebox[origin=c]{90}{\textbf{$T = 0.5$}}}
 & 6 SFT & $L_{KL}$ & 3.271 & 4.438 & 3.312 & 3.423 & 3.174 &  & 3.524 (0.002) & 3/1/7 \\
 & 6 SFT & $L_{KL} + 0.1L_{EAL}$ & 3.270 & 4.446 & 3.361 & 3.475 & 3.162 &  & 3.543 (0.022) & 3/1/7 \\
 & 3 SFT + 3 GRPO & \textcolor{blue}{$L_{KL} + 0.1L_{EAL} \to L_{KL} + 0.1L_{EAL} + 0.1L_{GRPO}$} & 3.313 & 4.460 & 3.372 & 3.505 & 3.221 &  & \textcolor{blue}{3.574} (0.021) & 3/1/7 \\
 & 6 SFT & \textcolor{orange}{$L_{KL} + 0.1L_{TV}$} & 3.332 & 4.443 & 3.297 & 3.490 & 3.175 &  & \textcolor{orange}{3.547} (0.006) & 3/1/7 \\
 & 6 SFT & \textcolor{red}{$L_{WTV}$} & 3.375 & 4.552 & 3.420 & 3.564 & 3.287 &  & \textcolor{red}{3.640} (0.014) & 3/1/7 \\
 & 6 SFT & $L_{W-WTV}$ &  &  &  &  &  &  &  &  \\
\midrule
\multirow{7}{*}{\rotatebox[origin=c]{90}{\textbf{$T = 1.0$}}}
 & 6 SFT & $L_{KL}$ & 3.266 & 4.420 & 3.146 & 3.786 & 3.316 &  & 3.587 (0.025) & 3/1/7 \\
 & 6 SFT & $L_{KL} + 0.1L_{EAL}$ & 3.290 & 4.480 & 3.203 & 3.858 & 3.369 &  & 3.640 (0.036) & 3/1/7 \\
 & 3 SFT + 3 GRPO & \textcolor{orange}{$L_{KL} + 0.1L_{EAL} \to L_{KL} + 0.1L_{EAL} + 0.1L_{GRPO}$} & 3.324 & 4.560 & 3.167 & 3.896 & 3.353 &  & \textcolor{orange}{3.660} (0.039) & 3/1/7 \\
 & 6 SFT & \textcolor{blue}{$L_{CE}$} & 3.242 & 4.462 & 3.370 & 3.920 & 3.399 &  & \textcolor{blue}{3.679} (0.091) & 3/1/7 \\
 & 6 SFT & $L_{LK}$ & 3.278 & 4.448 & 3.172 & 3.861 & 3.343 &  & 3.620 (0.029) & 3/1/7 \\
 & 6 SFT & $L_{KL} + 0.1L_{TV}$ & 3.282 & 4.526 & 3.131 & 3.869 & 3.366 &  & 3.635 (0.025) & 3/1/7 \\
 & 6 SFT & \textcolor{red}{$L_{WTV}$} & 3.378 & 4.599 & 3.468 & 4.048 & 3.526 &  & \textcolor{red}{3.804} (0.005) & 3/1/7 \\
 & 6 SFT & $L_{W-WTV}$ &  &  &  &  &  &  &  &  \\
\bottomrule
\end{tabular}%
}
\end{table}

Table~\ref{tab:sd3_merged} reports the acceptance length for Qwen3-32B/Qwen3-4B under tree
drafting. Under greedy verification, $L_{KL}+0.1L_{EAL}$ reaches $5.073$ against $5.057$
for $L_{KL}$ and the GRPO stage gives the best result at $5.088$, with $L_{LK}$ third at
$5.063$; $L_{WTV}$ is last at $5.039$. Under sampling verification $L_{WTV}$ leads by a
wide margin, reaching $5.496$ at $T=0.5$ against $4.975$ for $L_{KL}$, and $6.671$ at
$T=1.0$ against $5.960$. 

Table~\ref{tab:sd3_merged_chain} reports the same model pair under chain drafting. Under greedy verification, the ordering is consistent with the tree-based setting: the GRPO stage on top of $L_{KL}+0.1L_{EAL}$ achieves the highest acceptance length of $3.636$, followed by $L_{KL}+0.1L_{EAL}$ at $3.632$ and $L_{KL}$ at $3.615$, while $L_{WTV}$ performs worst at $3.603$. Under sampling verification, $L_{WTV}$ again achieves the highest acceptance length, reaching $3.640$ at $T=0.5$ and $3.804$ at $T=1.0$, compared with $3.524$ and $3.587$ for $L_{KL}$.

\begin{table}[!h]
\centering
\caption{Acceptance length on four benchmarks with Gemma4-31B as the target model and Gemma4-E2B as the draft model with chain-based speculative decoding under greedy verification and sampling temperature $T = 1.0$. Colored numbers indicate ranking by Avg.\ Acceptance Length within each setting: \textcolor{red}{red} denotes best.}
\label{tab:gemma_merged_chain}
\renewcommand{\arraystretch}{1.15}
\resizebox{0.9\textwidth}{!}{%
\begin{tabular}{lllccccc}
\toprule
\textbf{Setting} & Gemma4-31B & Gemma4-E2B & \multicolumn{5}{c}{\textbf{Acceptance Length}} \\
\cmidrule(lr){2-3} \cmidrule(lr){4-8}
 & \textbf{Stage} & \textbf{Loss} & \textbf{MT-Bench} & \textbf{GSM8K} & \textbf{HumanEval} & \textbf{Speech} & \textbf{Avg} \\
\midrule
\multirow{5}{*}{\rotatebox[origin=c]{90}{Greedy}}
 & 6 SFT  & $L_{KL}$                            & 4.897 & 5.984 & 5.795 & -- & 5.559 \\
 & 6 SFT  & $L_{CE}$                            & 4.811 & 5.873 & 5.743 & -- & 5.476 \\
 & 6 SFT  & $L_{LK}$                            & 4.905 & 5.996 & 5.797 & -- & 5.566 \\
 & 6 SFT  & \textcolor{red}{$L_{KL} + 0.1L_{EAL}$} & 4.913 & 6.009 & 5.800 & -- & \textcolor{red}{5.574} \\
 & 6 SFT  & $L_{WTV}$                           & 4.901 & 5.977 & 5.797 & -- & 5.558 \\
\midrule
\multirow{5}{*}{\rotatebox[origin=c]{90}{\textbf{$T = 1.0$}}}
 & 6 SFT  & $L_{KL}$                            & 5.084 & 6.078 & 5.936 & -- & 5.699 \\
 & 6 SFT  & $L_{CE}$                            & 5.102 & 6.084 & 5.957 & -- & 5.714 \\
 & 6 SFT  & $L_{LK}$                            & 5.095 & 6.091 & 5.944 & -- & 5.710 \\
 & 6 SFT  & $L_{KL} + 0.1L_{EAL}$              & 5.108 & 6.107 & 5.952 & -- & 5.722 \\
 & 6 SFT  & \textcolor{red}{$L_{WTV}$}          & 5.363 & 6.518 & 6.424 & -- & \textcolor{red}{6.102} \\
\bottomrule
\end{tabular}%
}
\end{table}

Table~\ref{tab:gemma_merged_chain} reports the acceptance length for
Gemma4-31B/Gemma4-E2B under chain drafting. Under greedy verification, $L_{EAL}$
again provides a small but consistent improvement over the $L_{KL}$ baseline, with
$L_{KL}+0.1L_{EAL}$ reaching $5.574$ against $5.559$. $L_{WTV}$ is competitive but
not better in this setting, at $5.558$. The advantage of WTV again emerges under
sampling verification. At $T=1$, $L_{WTV}$ achieves $6.102$ compared with $5.699$
for $L_{KL}$ and $5.722$ for $L_{KL}+0.1L_{EAL}$. 

\begin{table}[!h]
\centering
\caption{Acceptance length on four benchmarks with gpt-oss-120b as the target model and Qwen3-4B as the draft model with chain-based speculative decoding under greedy verification and sampling temperature $T = 1.0$. Colored numbers indicate ranking by Avg.\ Acceptance Length within each setting: \textcolor{red}{red} denotes best.}
\label{tab:gptoss_merged_chain}
\renewcommand{\arraystretch}{1.15}
\resizebox{0.9\textwidth}{!}{%
\begin{tabular}{lllccccc}
\toprule
\textbf{Setting} & gpt-oss-120b & Qwen3-4B & \multicolumn{5}{c}{\textbf{Acceptance Length}} \\
\cmidrule(lr){2-3} \cmidrule(lr){4-8}
 & \textbf{Stage} & \textbf{Loss} & \textbf{MT-Bench} & \textbf{GSM8K} & \textbf{HumanEval} & \textbf{Speech} & \textbf{Avg} \\
\midrule
\multirow{5}{*}{\rotatebox[origin=c]{90}{Greedy}}
 & 6 SFT  & $L_{KL}$                       & -- & -- & -- & -- & -- \\
 & 6 SFT  & $L_{CE}$                       & -- & -- & -- & -- & -- \\
 & 6 SFT  & $L_{LK}$                       & -- & -- & -- & -- & -- \\
 & 6 SFT  & $L_{KL} + 0.1L_{EAL}$         & -- & -- & -- & -- & -- \\
 & 6 SFT  & $L_{WTV}$                      & -- & -- & -- & -- & -- \\
\midrule
\multirow{5}{*}{\rotatebox[origin=c]{90}{\textbf{$T = 1.0$}}}
 & 6 SFT  & $L_{KL}$                       & -- & -- & -- & -- & -- \\
 & 6 SFT  & $L_{CE}$                       & -- & -- & -- & -- & -- \\
 & 6 SFT  & $L_{LK}$                       & -- & -- & -- & -- & -- \\
 & 6 SFT  & $L_{KL} + 0.1L_{EAL}$         & -- & -- & -- & -- & -- \\
 & 6 SFT  & $L_{WTV}$                      & -- & -- & -- & -- & -- \\
\bottomrule
\end{tabular}%
}
\end{table}

Table~\ref{tab:gptoss_merged_chain} reports the acceptance length for
gpt-oss-120b/Qwen3-4B under chain drafting. We evaluate the proposed objectives under
both greedy and sampling verification to examine whether the improvements from
acceptance-aware training generalize to a substantially larger target model.

\section{Conclusion}

In this paper we study the design of draft-model training losses that directly
target acceptance length in Speculative Decoding. We evaluate different losses across several
target--draft pairs, in both chain and tree drafting, and both greedy and sampling-based acceptance criteria. Under
greedy verification, adding $L_{EAL}$ on top of the KL anchor gives a consistent improvement over the $L_{KL}$ baseline, and the subsequent GRPO stage adds a
further gain. Under sampling verification, $L_{WTV}$ gives a substantially larger
improvement at both $T=0.5$ and $T=1.0$ in sampling-based speculative decoding. The two losses are therefore
complementary, each matching the acceptance criterion it was derived from.

\newpage

\bibliography{iclr2026_conference}
\bibliographystyle{iclr2026_conference}

\appendix

\clearpage
\section{Efficient On-Policy Training}

Compared with the training-time test (TTT) of EAGLE-3~\cite{li2025eagle3},
whose chain-per-position rollout is processed in parallel across positions
and implemented efficiently by SpecForge~\cite{li2026specforge}, our
training method poses two additional workload challenges. First, it
expands a draft tree rather than a chain at every position, which
multiplies the rollout workload and make attention mask complex. Second, its temporal variant rolls out
only at the realized draft window starts, each of which depends on the accepted length of the previous window, which introduces a sequential dependency across positions. We address these with a shared-prefix tree rollout kernel (Sec.~\ref{sec:tree_kernel}) and an estimator-driven walk with dynamic batching (Sec.~\ref{sec:temporal_batching}).

\subsection{Forest Rollout with Shared Prefix KV}
\label{sec:tree_kernel}

Tree rollout expands a draft tree at many start positions of the same
sequence, and all of these trees attend to the same teacher-forced draft
prefix. We therefore compute the prefix KV once and share it across all
starts, appending only each start's tree tokens, so that a batch of starts
forms a forest over a single shared prefix. Attention over this forest is
computed with FlashInfer~\cite{ye2025flashinfer}, with the tree ancestor
mask passed as a custom mask, so that all trees in the batch are expanded
together.

\begin{figure}[h]
  \centering
  \includegraphics[width=\textwidth]{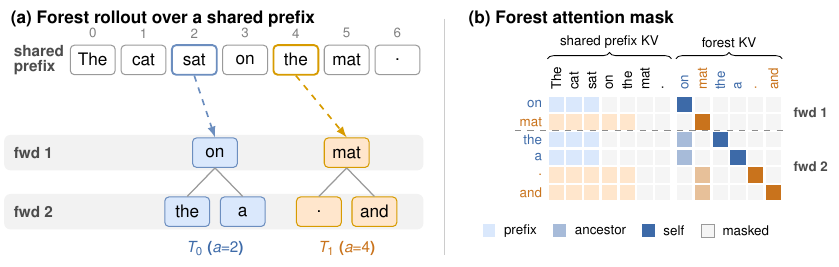}
  \caption{Forest rollout. (a) Draft trees at different anchors share one
  prefix KV, and nodes at the same depth across all trees are expanded in
  one forward pass. (b) The corresponding attention mask: each tree token
  attends to the prefix up to its anchor and to its ancestors in the forest.}
  \label{fig:forest_rollout}
\end{figure}

\subsection{Temporal Approximation and Dynamic Batching}
\label{sec:temporal_batching}

Training the temporal variant under greedy decoding requires the realized window starts, which make up only a fraction of about $1/\bar L)$ of the
positions, where $\bar L$ is the mean accepted length. There are two ways to obtain them:
\begin{itemize}
    \item \textbf{Roll out everywhere, replay at realized draft window starts.} Expand the draft
    tree at every position, walk the windows to find the starts, and
    backpropagate only through the rollouts at the starts. This keeps full
    position-level parallelism, but most of the rollout computation is
    discarded.
    \item \textbf{Roll out only at draft window starts.} Expand the draft tree only at
    the positions the walk actually visits. This saves computation in
    principle, which matters for compute-bound training, but introduces a
    sequential dependency: the next start is known only after the current
    window's accepted length is computed, which breaks parallelism across
    positions.
\end{itemize}
We follow the second option and recover parallelism through a temporal
approximation of the walk combined with dynamic batching, described below.

\paragraph{Approximation of the walked draft window} We first use a cheap
teacher forcing style estimator to find the possible draft windows start positive. At each position $t$, we take the draft's sorted top-$k$ log-probabilities $\ell_t^{(0)}\ge\ell_t^{(1)}\ge\cdots$ and the rank $r_t$ of the target's greedy token under the draft. We approximate the tree at $t$ by assuming that every branch continues under the ground-truth context, so that a path with per-depth ranks $(k_0,\ldots,k_{\gamma-1})$ scores
$\sum_{j}\ell_{t+j}^{(k_j)}$. Keeping the $K$ highest-scoring paths, the
approximate accepted length $\hat n_t$ is the longest prefix along which
some kept path satisfies $k_j=r_{t+j}$. We use $\hat n$ only to predict which positions the walk will visit and rollout each position in batch.

\paragraph{Approximation based temporal rollout}


Let $n_t$ denote the exact accepted length at position $t$, which is known
only after the draft tree at $t$ has been rolled out, and let $\hat n_t$
denote the estimator above, which is available at every position. The
exact walk $\mathcal{W}=\{s_1<s_2<\cdots\}$ starts at the first supervised
position $s_1$ and follows $s_{m+1}=s_m+n_{s_m}+1$.

Because each start depends on the previous one, we compute $\mathcal{W}$
over several parallel rounds, each of which rolls out one batch of draft trees. Let $\mathcal{E}_r$ be the set of positions rolled out after round $r$.
Starting from a position $h$, we simulate the walk as if the estimator
were exact: each window at $c$ is assumed to accept $\hat n_c$ tokens, so
the next one starts at $c+\hat n_c+1$. The positions visited form the
predicted chain
\begin{equation}
\mathcal{C}(h)=\{c_0,c_1,\ldots\},\qquad
c_0=h,\qquad c_{i+1}=c_i+\hat n_{c_i}+1,
\end{equation}
which we stop at the first position already rolled out or at the end of
the sequence.

The first round rolls out $\mathcal{E}_1=\mathcal{C}(s_1)$ in one batch.
After each round $r$, we advance the exact walk from $s_1$ as long as it
stays in $\mathcal{E}_r$. Along the predicted chain this continues as long
as $\hat n=n$; at the first chain position $c$ with $\hat n_c\neq n_c$,
the walk moves to $c+n_c+1$, which has not been rolled out. This position
is the frontier $h_r$, and round $r+1$ rolls out $\mathcal{C}(h_r)$,
setting $\mathcal{E}_{r+1}=\mathcal{E}_r\cup\mathcal{C}(h_r)$. We repeat
until the walk reaches the end of the sequence, which is shown in Figure \ref{fig:walk_rounds}'s left.

\paragraph{Dynamic batching}

\begin{wrapfigure}{r}{0.48\linewidth}
  \vspace{-12pt}
  \centering
  \includegraphics[width=\linewidth]{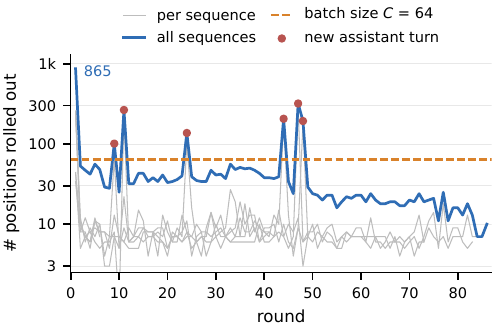}
  \vspace{-16pt}
  \caption{Positions rolled out per round without dynamic batching
  (6 ShareGPT sequences). Red: the walk enters a new assistant turn.}
  \label{fig:rollout_rounds}
  \vspace{-10pt}
\end{wrapfigure}

The estimator-driven walk suffers from a long tail of small rounds
(Fig.~\ref{fig:rollout_rounds}). The first round rolls out the predicted
walks of all sequences at once, but accounts for only about $20\%$ of all
rollouts. Every later round rolls out only the predicted chain from a
frontier left by an estimator error, a median of $7$ positions per
sequence, so even summed over sequences, most rounds stay below the
rollout batch size $C$.

To avoid this long tail, we fill every
round up to a multiple of $C$ with the positions most likely to be visited
next. From each walk frontier $h$, the next start is $h+\ell+1$ for some accepted
length $\ell\in\{0,\ldots,\gamma\}$. We estimate $P(n=\ell\mid\hat n)$ from
counts of $(\hat n_t,n_t)$ over the positions rolled out so far in the
micro-batch, smoothed with a prior that favors $n=\hat n$. Each candidate
position is scored by the probability of the branch that reaches it, i.e.,
the product of these terms along the way, and a best-first search adds the
most probable unexpanded positions until the round is full.

\begin{figure}[h]
  \centering
  \includegraphics[width=\textwidth]{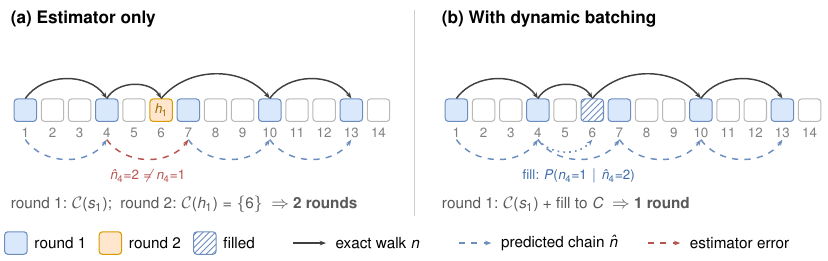}
\caption{Recovering the walk. Solid arcs: exact $n$; dashed arcs: estimator
$\hat n$. (a) An estimator error at position $4$ triggers a correction
round (orange). (b) Dynamic batching pre-fills the likely branch $6$, so
the walk completes in one round.}
  \label{fig:walk_rounds}
\end{figure}

\subsection{Results}
\label{sec:efficiency_results}

We profile training with a Qwen3-0.6B draft and a Qwen3-8B target on one
H200 with micro-batch size $8$ on ShareGPT, reporting the latency per micro-batch, averaged over $6$ micro-batches, and the
peak GPU memory. Table~\ref{tab:ablation} ablates the two components.

\paragraph{Shared prefix KV and batched tree kernel.}
Sharing the prefix KV across all starts keeps peak memory constant in the
rollout batch size $C$, at $86.9$\,GiB for $C\in\{128,256,512\}$, whereas
copying the prefix per start grows from $88.7$\,GiB at $C=128$ to
$122.6$\,GiB at $C=256$ and runs out of memory at $C=512$
(Table~\ref{tab:mem}). On its own, the kernel, run at $C=128$, reduces the
step time by $1.16\times$; its main benefit is to make large rounds
affordable, which dynamic batching then fills.

\paragraph{Approximation vs full rollout}

Table~\ref{tab:full_vs_walk} compares rolling out every position with rolling out only along the walk via approximator. The walk alone expands $58\%$ of the
positions but needs $89$ sequential rounds, and with the shared prefix kernel it is no faster than full rollout ($19.8$ vs.\ $19.3$\,s per
step): the saved computation is offset by the sequential dependency.
Dynamic batching resolves this, cutting the rounds to $25$ and the
expanded positions to $45\%$, and makes the walk $1.61\times$ faster than
the best full-rollout configuration ($11.5$ vs.\ $18.4$\,s).

\begin{table}[t]
\centering\small
\caption{Rolling out at every position (full) versus only along the walk,
with the shared prefix kernel. Expanded: fraction of the $11{,}843$
supervised positions per micro-batch. Top: matched chunk size $C$.
Bottom: best $C$ per method. Run-to-run noise is about $0.5$--$0.8$\,s.}
\label{tab:full_vs_walk}
\begin{tabular}{lccccc}
\toprule
Rollout & $C$ & Expanded & Rounds & Start search (s) & Step (s) \\
\midrule
\multicolumn{6}{l}{\emph{Matched $C$}} \\
full                    & 256 & 100\% & 1  & 17.33 & 19.32 \\
walk                    & 256 &  58\% & 89 & 17.79 & 19.80 \\
walk + dynamic batching & 256 &  55\% & 21 &  9.86 & 11.85 \\
\midrule
\multicolumn{6}{l}{\emph{Best $C$ per method}} \\
full                    & 512 & 100\% & 1  & 16.40 & 18.41 \\
walk                    & 256 &  58\% & 89 & 17.79 & 19.80 \\
walk + dynamic batching & 128 &  45\% & 25 &  9.46 & 11.46 \\
\bottomrule
\end{tabular}
\end{table}

\paragraph{Dynamic batching.}
Dynamic batching gives most of the gain, reducing the step time by
$1.59\times$ on its own. Combined with the kernel, the number of
sequential rounds drops from $89$ to $25$ and the step time from
$24.3$\,s to $11.4$\,s ($2.13\times$). The realized starts are unchanged,
and losses agree to within $4\times10^{-4}$. With both components, $C=128$
is the best setting; larger batches slightly increase the start search
time ($9.46$, $9.86$, and $11.40$\,s for $C=128$, $256$, and $512$).

\begin{table}[t]
\centering\small
\caption{Ablation of the tree rollout kernel and dynamic batching.}
\label{tab:ablation}
\begin{tabular}{llcccc}
\toprule
Prefix KV & Batching & $C$ & Start search (s) & Step (s) & Speedup \\
\midrule
per-start copy+SPDA & per round & 64  & 22.33 & 24.34 & $1.00\times$ \\
per-start copy+SPDA & dynamic   & 64  & 13.35 & 15.36 & $1.59\times$ \\
shared prefix+FlashInfer         & per round & 128 & 18.95 & 20.97 & $1.16\times$ \\
shared+FlashInfer         & dynamic   & 128 &  9.41 & 11.43 & $2.13\times$ \\
\bottomrule
\end{tabular}
\end{table}

\begin{table}[t]
\centering\small
\caption{Peak GPU memory of tree rollout.}
\label{tab:mem}
\begin{tabular}{lccc}
\toprule
Prefix KV & $C=128$ & $C=256$ & $C=512$ \\
\midrule
per-start copy & 88.7\,GiB & 122.6\,GiB & OOM \\
shared         & 86.9\,GiB &  86.9\,GiB & 86.9\,GiB \\
\bottomrule
\end{tabular}
\end{table}

\end{document}